\documentclass[fleqn,10pt, lineno]{wlpeerj}

\usepackage{amssymb}
\usepackage{listings}
\usepackage{xcolor}
\usepackage{graphicx}
\usepackage{subcaption}
\usepackage{makecell}
\usepackage{titlesec}

\definecolor{codegreen}{rgb}{0,0.6,0}
\definecolor{codegray}{rgb}{0.5,0.5,0.5}
\definecolor{codepurple}{rgb}{0.58,0,0.82}
\definecolor{backcolour}{rgb}{0.95,0.95,0.92}

\lstdefinestyle{mystyle}{
 backgroundcolor=\color{backcolour},  commentstyle=\color{codegreen},
 keywordstyle=\color{magenta},
 numberstyle=\tiny\color{codegray},
 stringstyle=\color{codepurple},
 basicstyle=\ttfamily\footnotesize,
 breakatwhitespace=false,     
 breaklines=true,         
 captionpos=b,          
 keepspaces=true,         
 numbers=left,          
 numbersep=5pt,         
 showspaces=false,        
 showstringspaces=false,
 showtabs=false,         
 tabsize=2
}

\title{Linguacodus: A Synergistic Framework for Transformative Code Generation in Machine Learning Pipelines}

\author[1]{Ekaterina Trofimova}
\author[1]{Emil Sataev}
\author[2,3]{Andrey Ustyuzhanin}
\affil[1]{HSE University, Moscow, Russia}
\affil[2]{Constructor University, Bremen, Germany}
\affil[3]{Institute for Functional Intelligent Materials, National University of Singapore, Singapore
}
\corrauthor[1]{Ekaterina Trofimova}{etrofimova@hse.ru}

\begin{abstract}
In the ever-evolving landscape of machine learning, seamless translation of natural language descriptions into executable code remains a formidable challenge. This paper introduces Linguacodus, an innovative framework designed to tackle this challenge by deploying a dynamic pipeline that iteratively transforms natural language task descriptions into code through high-level data-shaping instructions.
The core of Linguacodus is a fine-tuned large language model, empowered to evaluate diverse solutions for various problems and select the most fitting one for a given task. This paper details the fine-tuning process and sheds light on how natural language descriptions can be translated into functional code. Linguacodus represents a substantial leap towards automated code generation, effectively bridging the gap between task descriptions and executable code. It holds great promise for advancing machine learning applications across diverse domains. Additionally, we propose an algorithm capable of transforming a natural description of an ML task into code with minimal human interaction. In extensive experiments on a vast machine learning code dataset originating from Kaggle, we showcase the effectiveness of Linguacodus. The investigations highlight its potential applications across diverse domains, emphasizing its impact on applied machine learning in various scientific fields.
\end{abstract}

\begin{document}

\flushbottom
\maketitle
\thispagestyle{empty}

\section*{Introduction}
\label{sec:introduction}
Automated code generation from natural language, a field often referred to as natural language programming (NLP), holds the promise of simplifying programming tasks and enhancing the software development process~\citep{lei2013natural},~\citep{desai2016program},~\citep{wang2023natural}, particularly in the field of machine learning (ML)~\citep{chandel2022training}. The demand for efficient ML solutions is continuously rising, showcasing the significance of this technology in streamlining programming tasks and enhancing software development processes. ML has transformed human lives and significantly impacted scientific research and engineering~\citep{alpaydin2021machine}. It has emerged as a standard tool in various domains, revolutionizing the way tasks are approached and problems are solved~\citep{jung2022machine}. With the increasing reliance on ML solutions, the ability to swiftly and accurately translate ambiguous task descriptions into functional code has become increasingly vital.

Early endeavors in code generation from natural language primarily rely on rule-based systems and template-based approaches~\citep{gulwani2017program}. These methods suffer from limited expressiveness and scalability as they struggle to accommodate the variability and complexity of human and coding languages~\citep{allamanis2018survey}.

\cite{vaswani2017attention} introduce the Transformer architecture, a cornerstone in many natural language processing tasks, including code generation. Transformer-based models excel in capturing long-range dependencies and contextual information, leading to significant improvements in code generation quality. The synergy of deep learning techniques and the availability of extensive training data has transformed the landscape of code generation from natural language~\citep{vaithilingam2022expectation}, paving the way for the development of Large Language Models (LLMs). These LLMs exhibit the capability to learn intricate mappings between textual inputs and executable code.

While significant progress has been made in code generation from natural language, there remains a substantial gap in effectively transforming complex machine learning task descriptions into precise, executable code~\citep{yin2022natural},~\citep{wen2024grounding}. Current generative models often produce common yet suboptimal code snippets based on textual input, failing to capture the nuanced requirements of specific ML tasks. This gap exists primarily due to the complexity and variability of ML tasks, which often require domain-specific knowledge and customized approaches. The challenge also lies in converting detailed ML task narratives into a structured series of code components, as LLMs excel more with direct instructions. By "instructions" we mean the high-level guidance provided to the model for generating specific outputs (see Methodology Section).  Moreover, the difficulty is in maintaining coherence and logical flow throughout longer code sequences necessary for complete ML solutions. Addressing this knowledge gap can accelerate the development and prototyping of ML solutions, democratize ML development, and enhance the reproducibility and standardization of ML research.

\begin{figure}
\begin{center}
\includegraphics[trim={7cm 9cm 7cm 9cm},clip, width=1\linewidth]{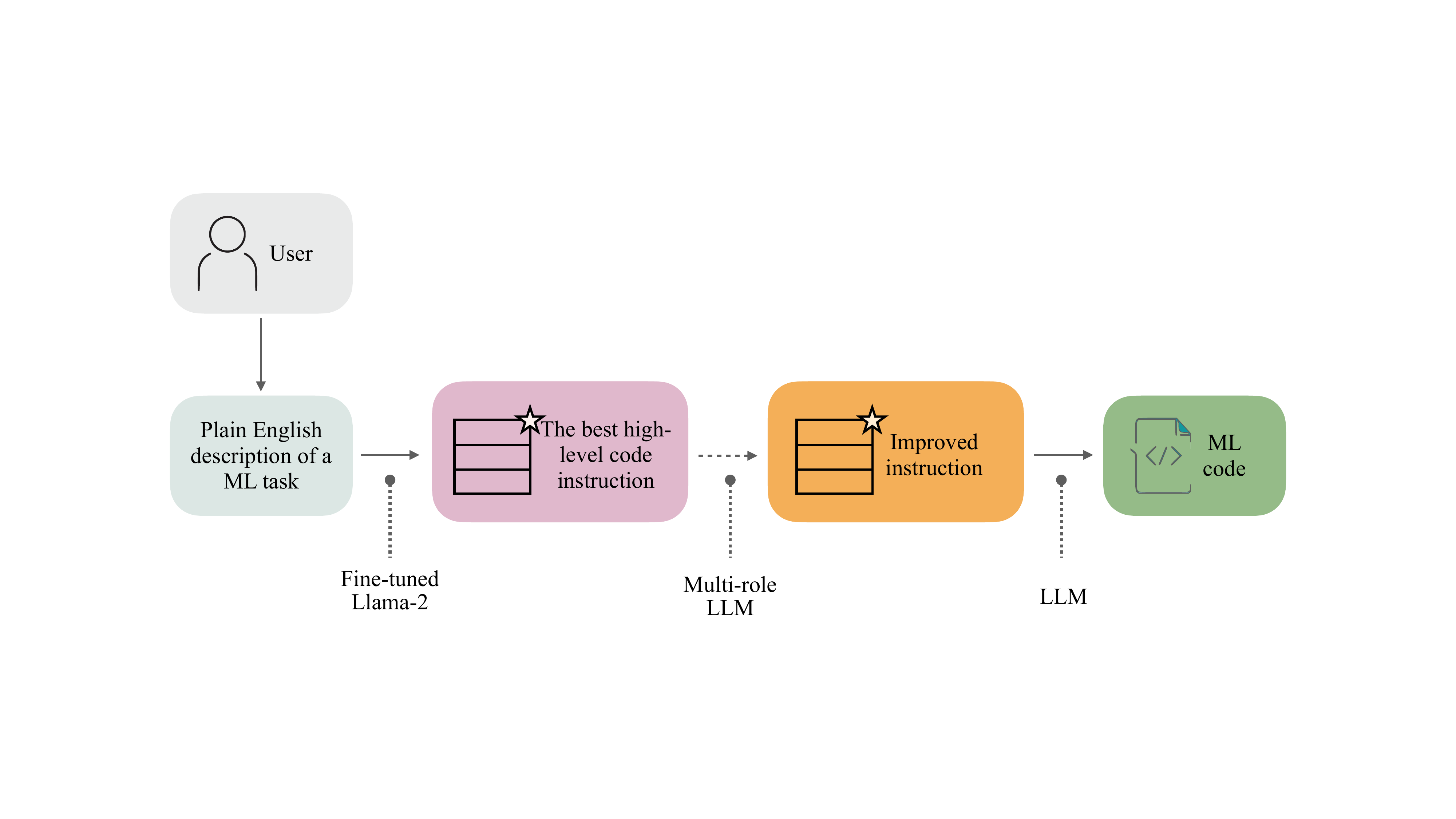}
\end{center}
\caption{Linguacodus takes in the user-provided description of a machine learning task and generates an optimal solution instruction. This instruction is then optionally refined using Multi-role LLM. Another LLM is employed to infer executable ML code based on the enhanced instruction. The resulting code represents the most effective solution for the specified task.}
\label{fig:inference}
\end{figure}

Our approach, Linguacodus, seeks a more accurate and flexible solution. It involves a two-step process: first, it transforms the human-provided ML task descriptions into explicit, high-level instructions. This step ensures the instructions are clear, verifiable, and understandable to the user, laying a solid foundation for the next phase. Then, these high-level instructions are translated into machine-compilable code representation, specifically Python code in our case, with the potential for extension to other programming languages (Fig.~\ref{fig:inference}). This method not only accommodates the intricate nature of ML tasks but also enhances the control and precision in the code generation process, meeting the need for understanding and controlled production of code in ML applications.

By converting human language into executable code, Linguacodus enables quick prototyping, ease iteration, and facilitate the deployment of ML models, potentially democratizing software development. This breakthrough allows individuals without extensive coding skills to engage in creating complex ML tasks, promoting innovation across various disciplines. The drive for such technology underlines a vision to broaden ML's reach and impact, simplifying the development process and inviting a wider audience to contribute to technological advancements. Portions of this text were previously published as part of a preprint~\citep{trofimova2024linguacodus}.

Our main contributions can be summarized as follows:

\textit{A Controllable Transformation Framework:} We present a framework for the controlled transformation of ML task descriptions into solution instructions, involving fine-tuning the Llama 2 model using pairs of ML task descriptions and instructions. 

\textit{Instruction-Based Sequential Generation:} We demonstrate the efficacy of executing instructions for sequential generation, producing compilable code with promising results based on evaluation metrics.

The rest of the paper is organised as follows. Section 'Related Work' explores the application of Large Language Models (LLMs) in code generation, text-to-code conversion, controllable generation, and automating problem-solving tasks, shedding light on the limitations of LLMs in ML code synthesis. 
Section 'Methodology' provides an overview of the Linguacodus framework. Section 'Experimantal Results and Analysis' describes the experiments and validation of the approach, highlighting the effectiveness of Linguacodus in transforming plain English descriptions of ML tasks into executable code. Sections 'Discussion' and 'Limitations' discusses and critically examines the limitations our approach. 'Future work' suggest the future perspectives of the work. Finally, Section 'Conclusion' summarizes and concludes the paper.

\section*{Related Work}
\label{sec:related}

Code generation from developer's requirements has emerged as a compelling area of research, bridging the realms of NLP and programming languages~\citep{liu2020deep}. Traditional methodologies for code synthesis from human language have historically leaned on formal semantic representations of natural language ~\citep{winograd1972understanding},~\citep{harel1990statemate},~\citep{buse2012synthesizing}. However, formal specifications require manual creation and maintenance, making them labor-intensive and difficult to scale for large codebases or complex systems~\citep{raychev2014code}.

\cite{ling2016latent} automatically predict code snippets directly from natural language inputs by proposing Latent Predictor Networks (LPN). LPN encapsulates the latent variable model for capturing the underlying structure of the input natural descriptions, and the predictor network for mapping the latent representations to corresponding code snippets.

Meanwhile,~\cite{rabinovich2017abstract},~\cite{yin2017syntactic} and~\cite{yin2018tranx} emphasize the importance of incorporating syntax awareness into the neural network architectures. The researchers leverage the Abstract Syntax Tree to capture the well-defined structure in the target programming syntax. Additionally, Long Short-Term Memory (LSTM) networks are employed to capture long dependencies in natural language sequences. However, these methods predominantly rely on a single NL statement.

In contrast,~\cite{agashe2019juice} tackle the task of interactive general-purpose code generation by incorporating a full sequence of previous natural language and code blocks as context within a Python Jupyter notebook environment~\citep{kluyver2016jupyter}. Still, the work is limited to the domain defined by the JuICe dataset, consisting of code snippets and corresponding markdowns, and does not utilize general task descriptions as inputs for code generation.

Utilizing vast amounts of code and natural language data has been made possible through pre-training techniques~\citep{radford2018improving}, ~\citep{devlin2018bert}. By leveraging pre-trained models, like CodeBERT~\citep{feng2020codebert}, researchers strive to capture comprehensive representations of both code and language semantics. This enables the models to produce code from natural language descriptions that are not only more accurate but also contextually relevant. Such models offer versatility in code-related tasks, including code generation, summarization, and recommendation.


CoditT5 \citep{zhang2022coditt5} is another language model that generates edit-based output sequences from corrupted input sequences. Models like CoditT5 enhance code generation capabilities, aligning them more closely with user requirements.

Modern code generation approaches often rely on general-purpose transformers, exemplified by GPT-3. Codex~\citep{Chen2021EvaluatingLL}, a notable model in this category, showcases the potential to generate code snippets directly from natural language prompts. AlphaCode~\citep{doi:10.1126/science.abq1158} extends this foundation, emphasizing the significance of code diversity and improving contextual understanding in LLMs.

In parallel, text-to-code conversion has gained prominence. PaLM-Coder~\citep{chowdhery2023palm} presents a method for converting natural language descriptions into code, focusing on Java code generation. OpenAI models~\citep{openai2023gpt4},~\citep{bubeck2023sparks} have further extended the capabilities of LLMs in understanding and generating code from textual prompts.

Controllable code generation is an emerging subfield with significant potential. CTRL~\citep{keskar2019ctrl} is a conditional language model for controlled code generation. The model focuses on allowing users to specify conditions that influence the generated code, providing a level of control over the output. Texygen \citep{song2018texygen} is a benchmarking platform for evaluating text generation models, including those designed for code generation. This platform facilitates the assessment of controllable code generation models by offering standardized evaluation metrics and tasks.

In automating problem-solving tasks, researchers have actively explored solutions such as AutoGluonTabular~\citep{erickson2020autogluontabular} and H2O AutoML~\citep{LeDell2020H2OAS}. These frameworks offer automated machine learning capabilities to streamline the model development process and improve prediction accuracy.

In particular, LightAutoML~\citep{vakhrushev2022lightautoml} tailors itself to the distinctive needs of large financial services companies companies. It provides solutions for handling large datasets with diverse types, non-stationary data, and specific validations, making it well-suited for complex financial analysis tasks. 

Another recent AutoML framework, HuggingGPT~\citep{shen2023hugginggpt}, utilizes ChatGPT for task planning, model selection, subtask execution, and result summarization. HuggingGPT demonstrates versatility across a wide range of AI tasks, including natural language understanding and automated problem-solving.

\cite{nair2023dera} present the dialog-enabled resolving agents (DERA), aiming for accurate output generation. DERA enhances the conversational abilities of LLMs by incorporating two distinct agent types: a Researcher, responsible for processing information and identifying critical problem components, and a Decider, capable of autonomously integrating the Researcher's information and making judgments on the final output. Although the DERA paradigm was initially used in healthcare, one can notice the potential applicability of multi-agent LLM in various training fields.

While automated machine learning offers structured workflow optimization, ML code generation based on natural language descriptions provides seamless integration into existing systems and customization for domain-specific tasks.

The recent advancements in code generation driven by LLMs have witnessed notable progress. Thus, OpenAI GPT models~\citep{openai2023gpt4},~\citep{bubeck2023sparks}, although not explicitly designed for code generation, have demonstrated proficiency in generating code snippets and understanding programming-related prompts. The generative capabilities of GPT models make them versatile tools for interpreting and translating natural language descriptions into executable code.

Google's PaLM 2~\citep{anil2023palm} undergoes pre-training on a vast dataset encompassing web pages and source code, making it valuable for code debugging, completion, and generation across multiple programming languages. The model's dual focus on semantic parsing and language model pre-training enhances its ability to comprehend and generate code based on diverse natural language inputs.

One of the leading publicly available LLMs for code generation is Code Llama \citep{rozière2024code}. An extension of Llama 2 \citep{touvron2023llama}, Code Llama comes in two variations: a code producer and its instruction-specific refinement, Code Llama - Instruct. Code Llama - Instruct surpasses Code Llama in providing more helpful and secure responses in natural language, ensuring a more dependable performance. However, the generated instructions are generally broad-purpose and lack easy assessability regarding their suitability for specific tasks.

While OpenAI's ChatGPT and similar LLMs have demonstrated remarkable capabilities in various natural language understanding tasks, they do have some inherent limitations in the context of ML code generation:
\begin{enumerate}
  \item Lack of Specificity: LLMs often generate code snippets that lack specificity for specific ML tasks. The generated code may be overly general and not finely tailored to the requirements of complex machine learning workflows.

  \item Limited Control Over Code Generation: Users have limited control over the fine-tuning process of LLMs, making it challenging to enforce specific guidelines or constraints during the generation of ML code. This lack of control may result in variations in code quality and suitability for different tasks.

  \item Handling Ambiguity: Natural language descriptions of ML tasks can be inherently ambiguous. LLMs may struggle to disambiguate between multiple potential interpretations, leading to code snippets that may not accurately capture the intended meaning of the task.

  \item Inability to Learn Task-Specific Patterns: While proficient in learning patterns from diverse data, LLMs may face challenges in capturing task-specific patterns relevant to ML code generation. This limitation can result in generated code that lacks the specificity required for specialized tasks.

  \item Evaluation Metrics and Validation: The evaluation metrics for assessing the quality of generated code may not always align with the specific requirements of ML tasks. LLMs may prioritize generating syntactically correct code without necessarily ensuring the semantic correctness or optimization of the generated solutions.

\end{enumerate}

\begin{figure}[h]
\begin{center}
\includegraphics[trim={0cm 0cm 0cm 0cm},clip, width=1\linewidth]{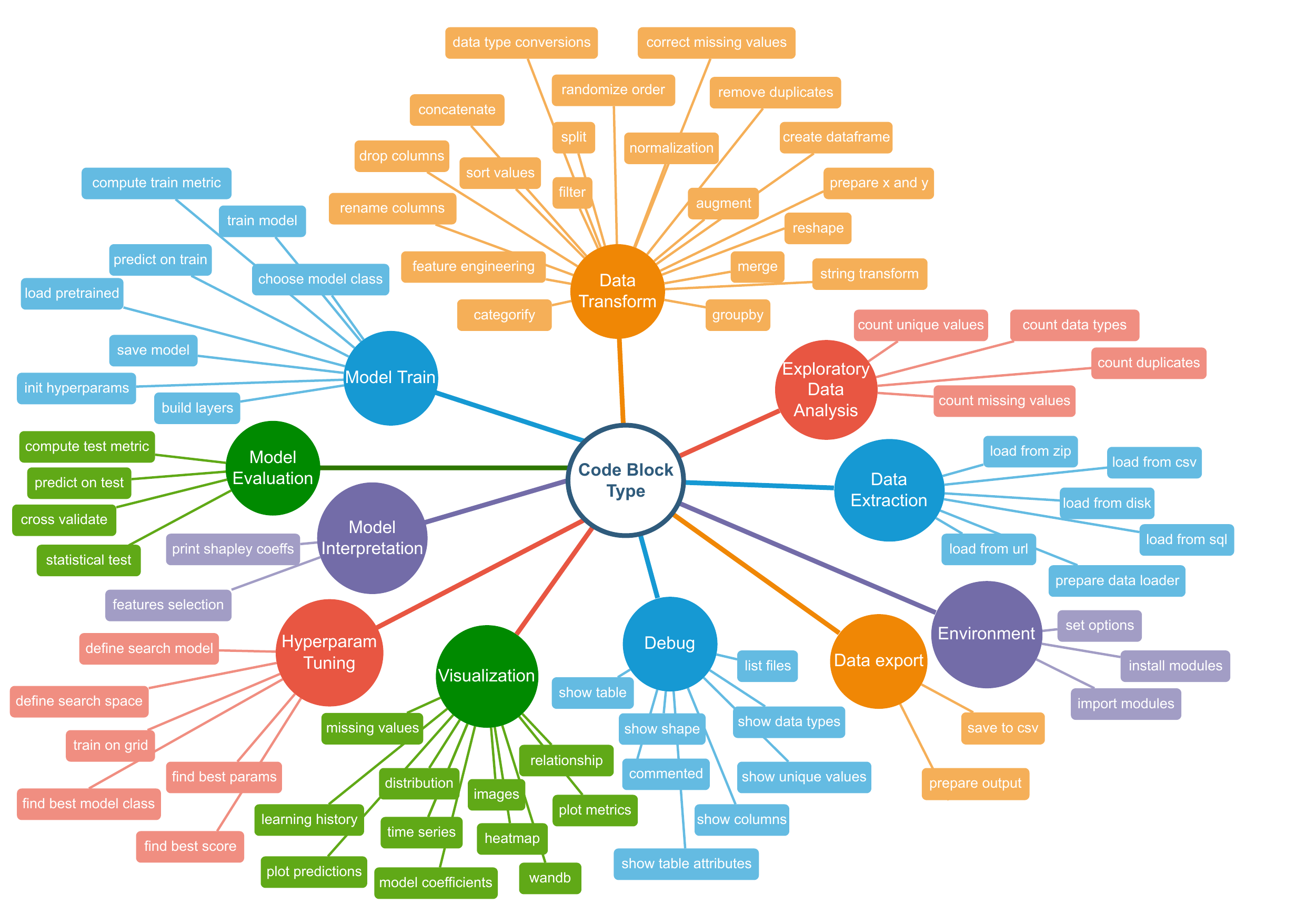}
\end{center}
\caption{Code4ML taxonomy tree. Reproduced from~\citep{code4ml}, with permission of the authors.}
\label{fig:code4ml}
\end{figure}

Addressing these challenges requires a hybrid approach involving specialized ML code datasets and dimensional reduction within the learning space for LLM fine-tuning. The Code4ML~\citep{code4ml} is a comprehensive corpus comprising of a) Kaggle challenge descriptions in natural language, b) Jupyter notebooks and their scores, c) Python code snippets, and d) competition-related metadata. This metadata includes formal descriptions of challenge datasets and scoring metrics. Code4ML relies on a knowledge taxonomy tree (Fig.~\ref{fig:code4ml}) to categorize various Jupyter notebook code snippets. A description of a challenge solution in terms of the classes of this taxonomy significantly reduces the dimensionality of a code generation problem compared to the direct generation of code by using task description as a prompt. However, Code4ML lacks annotation for all code snippets. This limitation is addressed through the taxonomy-based categorization introduced by \citep{code4mlclass}.

\section*{Methodology}
\label{sec:approach}

\begin{figure*}[h]
\begin{center}
\includegraphics[trim={0cm 8cm 0cm 8cm},clip, width=1\linewidth]{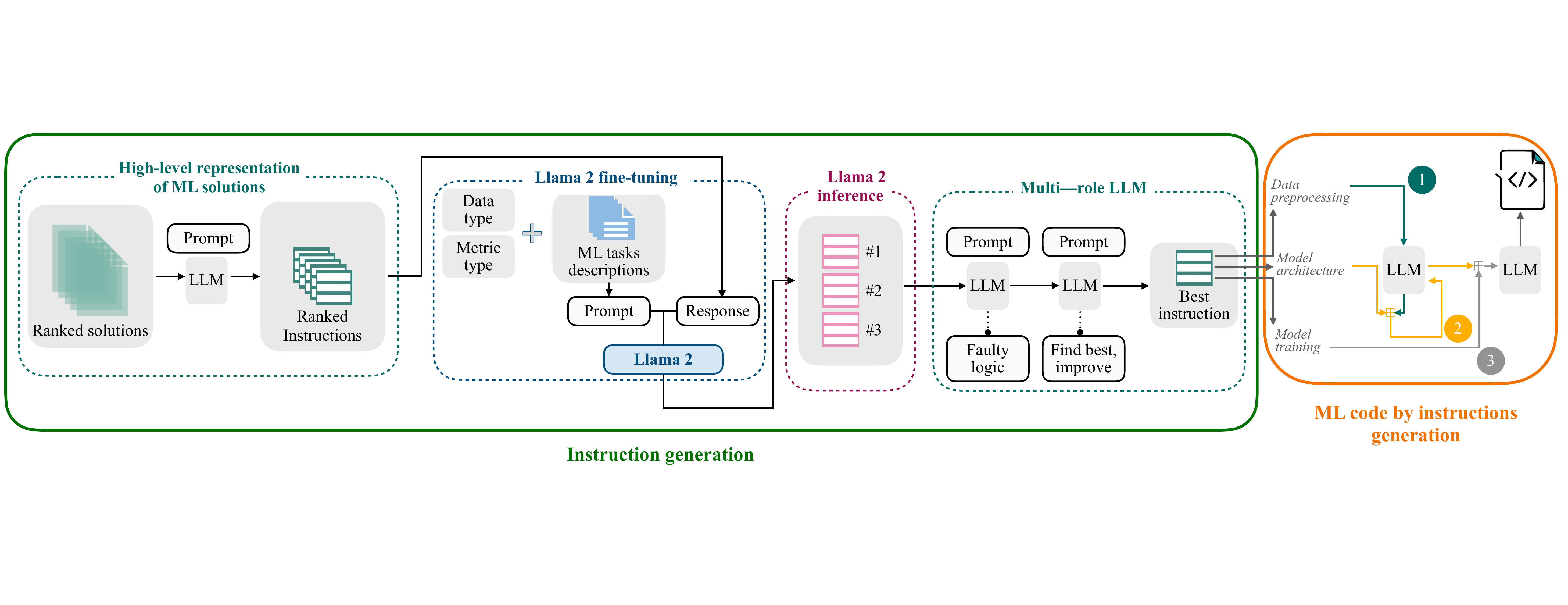}
\end{center}
\caption{Overall Linguacodus training framework.}
\label{fig:instructions}
\end{figure*}

This section presents a comprehensive overview of the Linguacodus. Fig.~\ref{fig:instructions} depicts the two stages of the framework. Initially, utilizing the fine-tuned Llama 2, we generate the most appropriate instruction, encapsulating the high-level core information of a generalized ML solution, tailored to a specific ML task. Subsequently, this instruction undergoes a sequential transformation into programming code through prompts with GPT-3.5.

\subsection*{Instruction Creation} 
\label{subsec:instructions_gen}

\begin{figure}[h!]
  \begin{minipage}[h]{0.48\textwidth}
\center{\includegraphics[width=1\linewidth]{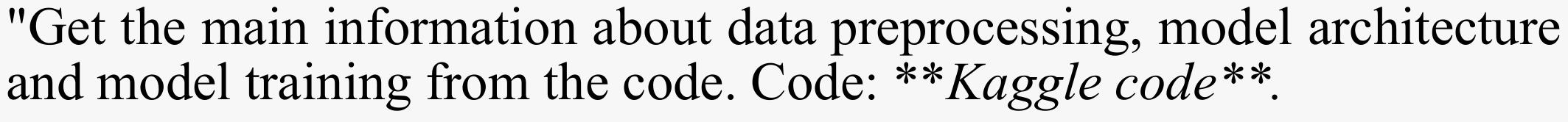}}\\
\caption{Prompt for ML instructions retrieving.}
\label{fig:prompt_1}
  \end{minipage}\hfill
  \begin{minipage}[h]{0.48\textwidth}
\center{\includegraphics[width=1\linewidth]{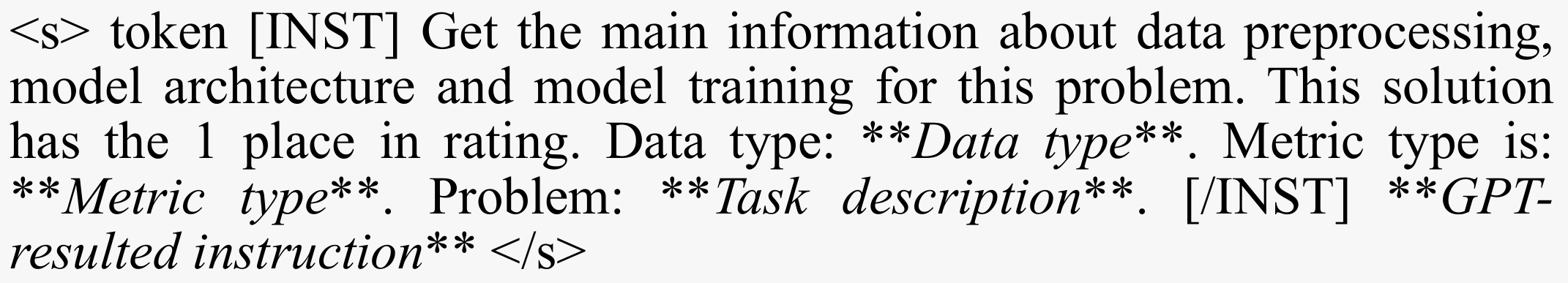}}\\
\caption{Llama 2 fine-tune input.}
\label{fig:prompt_2}
  \end{minipage}\hfill
\end{figure}

To extract the high-level code instructions, we've devised a four-stage framework:

1. \textit{High-Level Solution Representation}: We begin by creating high-level representations of ML solutions. To refine the quality of our dataset, the solutions undergo a ranking process based on their scores. Each solution is intricately linked to the natural language description of the ML task. Linguacodus utilizes the LLM to extract critical information regarding data preprocessing, model architecture, and the training procedure from existing code solutions. This information forms the high-level ML instruction.  Fig.~\ref{fig:prompt_1} illustrates the precise input prompt presented to the model.


2. \textit{Llama 2 Fine-Tuning}: Then, we utilize the acquired instructions as inputs for fine-tuning the open-source Llama 2 7b model. To ensure the relevance of the instructions to the machine learning (ML) task, we leverage the original code's quality evaluation in the form of a score. The retrieved instructions are ranked based on their significance to the ML task. Furthermore, we furnish the Llama 2 model with essential information presented as prompts, including the task description, metric details, and data type information. The prompt-completion pair used in this stage is visually depicted in Fig.~\ref{fig:prompt_2}, with the separation marked by the [/INST] token. This comprehensive approach enhances the fine-tuning process, incorporating the quality ranking of instructions and pertinent task details for optimal model adaptation. Llama models have been pre-trained on vast amounts of data. By fine-tuning, we leverage this extensive knowledge and adapt it to specific tasks, often achieving state-of-the-art results with less data and time. The fine-tuning details are summarised in Appendix~\ref{App A}.

\begin{figure}[h!]
  \begin{minipage}[h]{0.48\textwidth}
\center{\includegraphics[width=1\linewidth]{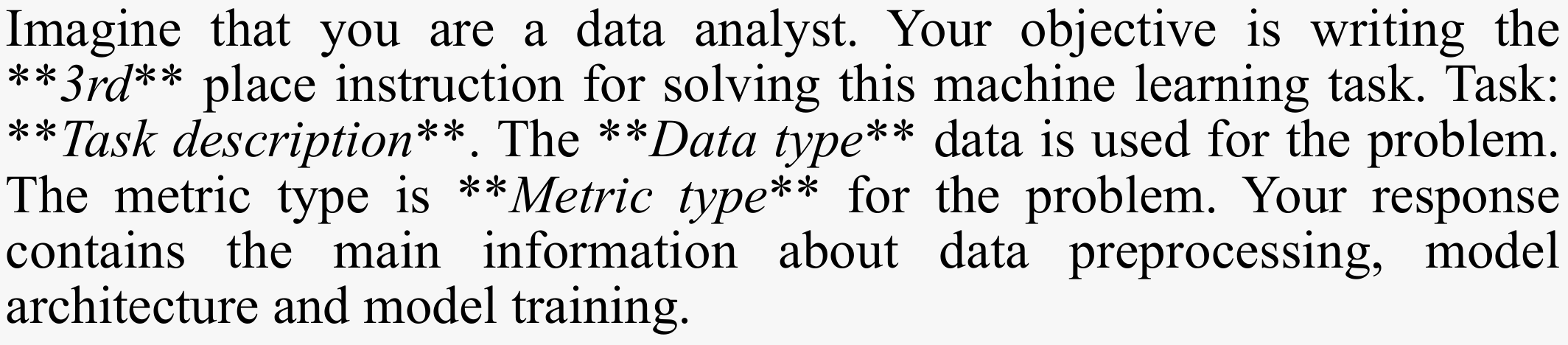}}\\
\caption{Prompt for Llama 2 inference.}
\label{fig:prompt_3}
  \end{minipage}\hfill
\end{figure}

3. \textit{Llama 2 Inference}: Next, we infer Llama 2 to select the top 3 most valuable instructions by specifying their rank using a dedicated prompt, as shown in Fig.~\ref{fig:prompt_3}. 

\begin{figure}[h!]
  \begin{minipage}[h]{0.48\textwidth}
\center{\includegraphics[width=1\linewidth]{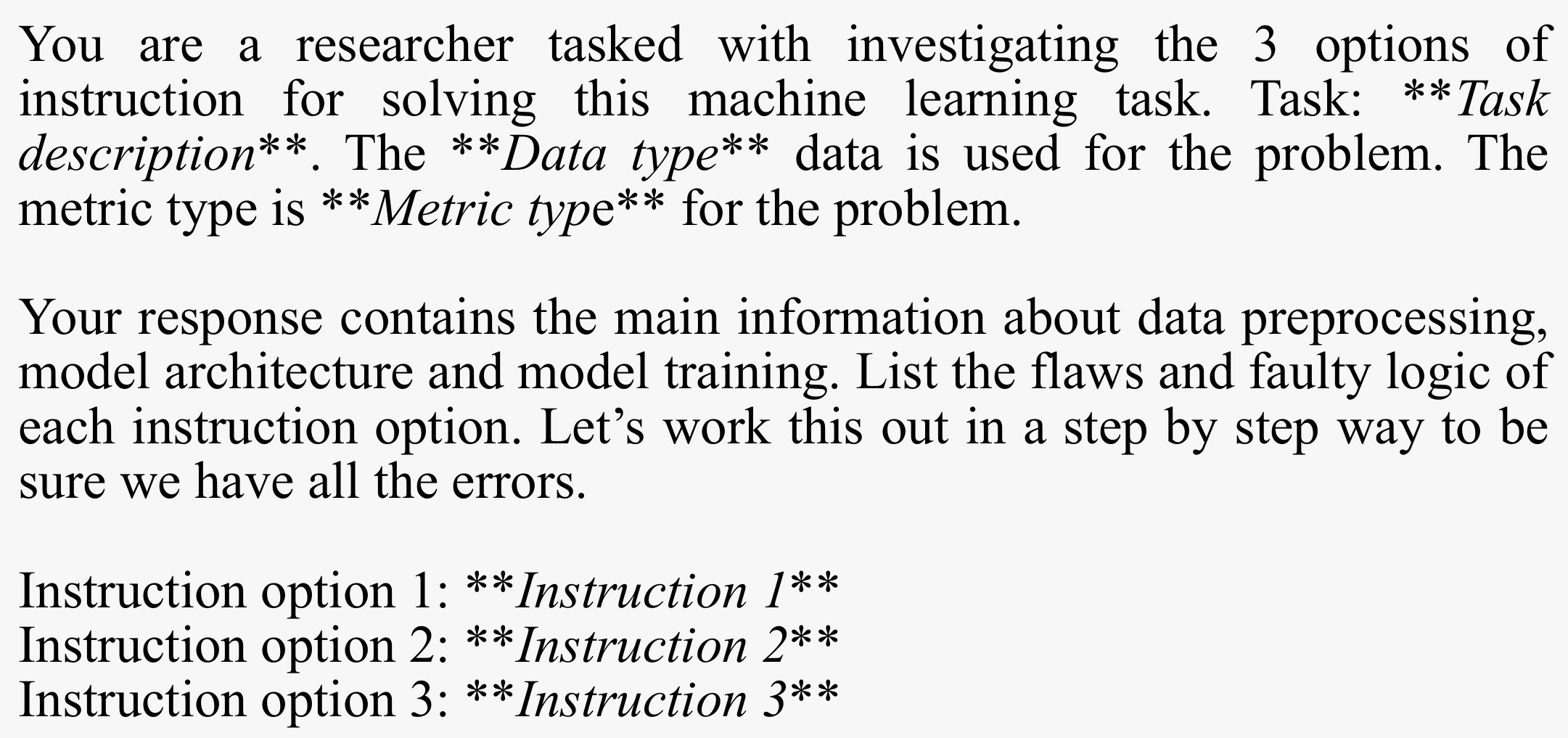}}\\
\caption{Prompt for Multi-agent LLM for best instruction choice.}
\label{fig:prompt_4}
  \end{minipage}\hfill
  \begin{minipage}[h]{0.48\textwidth}
\center{\includegraphics[width=1\linewidth]{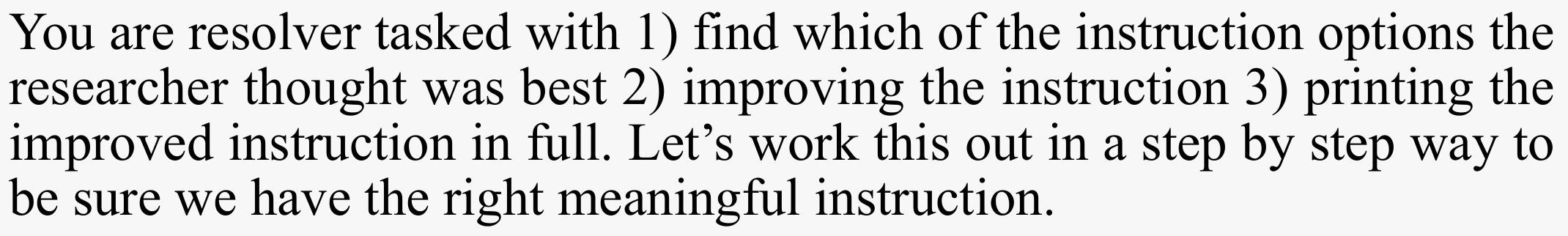}}\\
\caption{Prompt for Multi-agent LLM for best instruction improvement.}
\label{fig:prompt_4_}
  \end{minipage}\hfill
\end{figure}

4. \textit{Iterative enhancing LLM responses through multi-agent LLM}: The inferred instructions then undergo further refinement with the assistance of multi-agent LLM. 
The primary goal of multi-agent LLM is to identify any logical errors in the provided instructions and subsequently choose the best option from the three variants, thereby enhancing the overall quality of the instructions. This intelligent processing is elucidated in Fig.~\ref{fig:prompt_4}, and~\ref{fig:prompt_4_}.

\subsection*{ML Code by Instruction Generation}
The second stage of our approach centers on the actual code generation, building upon the instructions obtained in the previous step. In this phase, we harness the capabilities of language models to transform these instructions into functional and well-structured code that aligns with the underlying ML tasks.

\begin{figure*}[h]
\begin{center}
\includegraphics[trim={0cm 0cm 0cm 0cm},clip, width=1\linewidth]{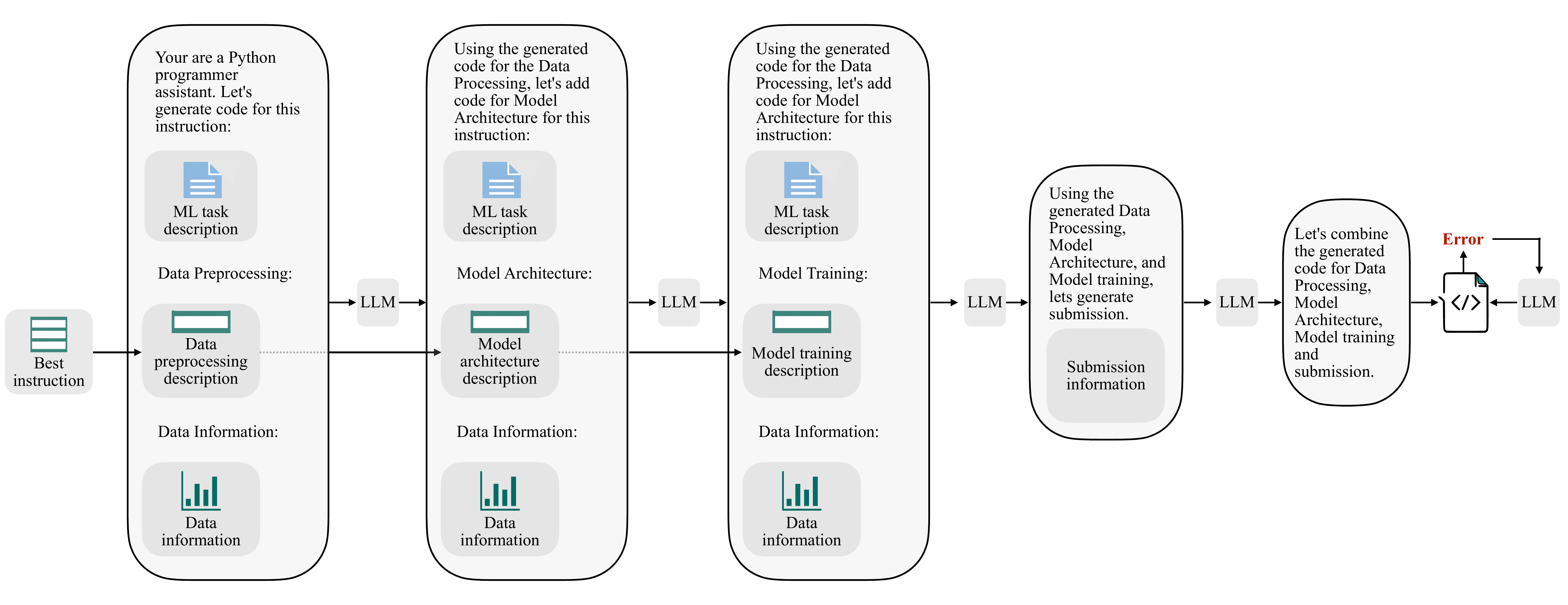}
\end{center}
\caption{Linguacodus instruction to code sequential transformation scheme.}
\label{fig:code}
\end{figure*}

Fig.~\ref{fig:code} precisely represents the sequential pipeline involved in the instruction-to-code transformation. We have separated the code synthesis into the stages of Data Preprocessing, Model Architecture, and Model Training. Additionally, we have also introduced a submission block to enable the testing of results on the Kaggle platform. The next step in this pipeline involves integrating all the generated code segments. To mitigate the possible execution problems, Linguacodus employs an error-fixing procedure, running it up to three times. In this process, the same LLM agent, responsible for integrating all code components iteratively, inputs the errors without any additional specifications.

This phase forms the critical bridge between the high-level ML instructions and the executable code, ensuring that the generated code adheres to the provided instructions and produces practical solutions for the intended ML tasks.

\section*{Experimental results and analysis}
\label{sec:experiments}

\subsection*{Dataset}

Our research relies on the Code4ML dataset, focusing on Kaggle competitions encompassing all metric categories except 'points,' 'significance,' and 'custom loss.' We curate the top 75 solutions for retrieving high-level instructions from these competitions. It is essential to highlight that specific contests may have fewer than 75 solutions available for selection.

\begin{figure}[h!]
  \begin{minipage}[h]{0.48\textwidth}
\center{\includegraphics[trim={1cm 1cm 1cm 1cm},clip, width=1\linewidth]{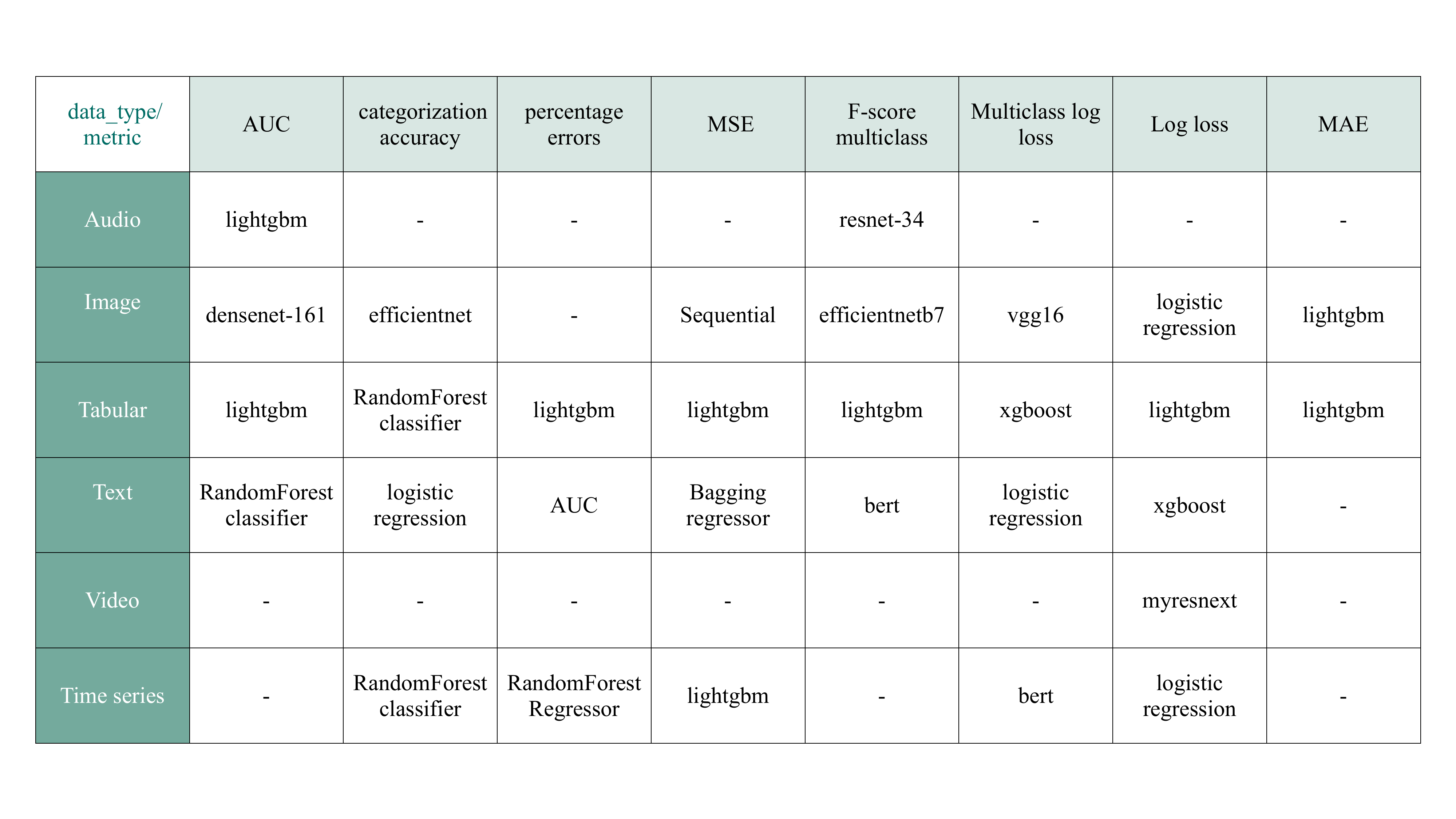}}\\
\caption{The most popular model choice among retrieved Kaggle solutions based on metric and data type.}
\label{fig:data}
  \end{minipage}\hfill
  \begin{minipage}[h]{0.48\textwidth}
\center{\includegraphics[trim={0cm 2cm 0cm 0cm},clip,width=1\linewidth]{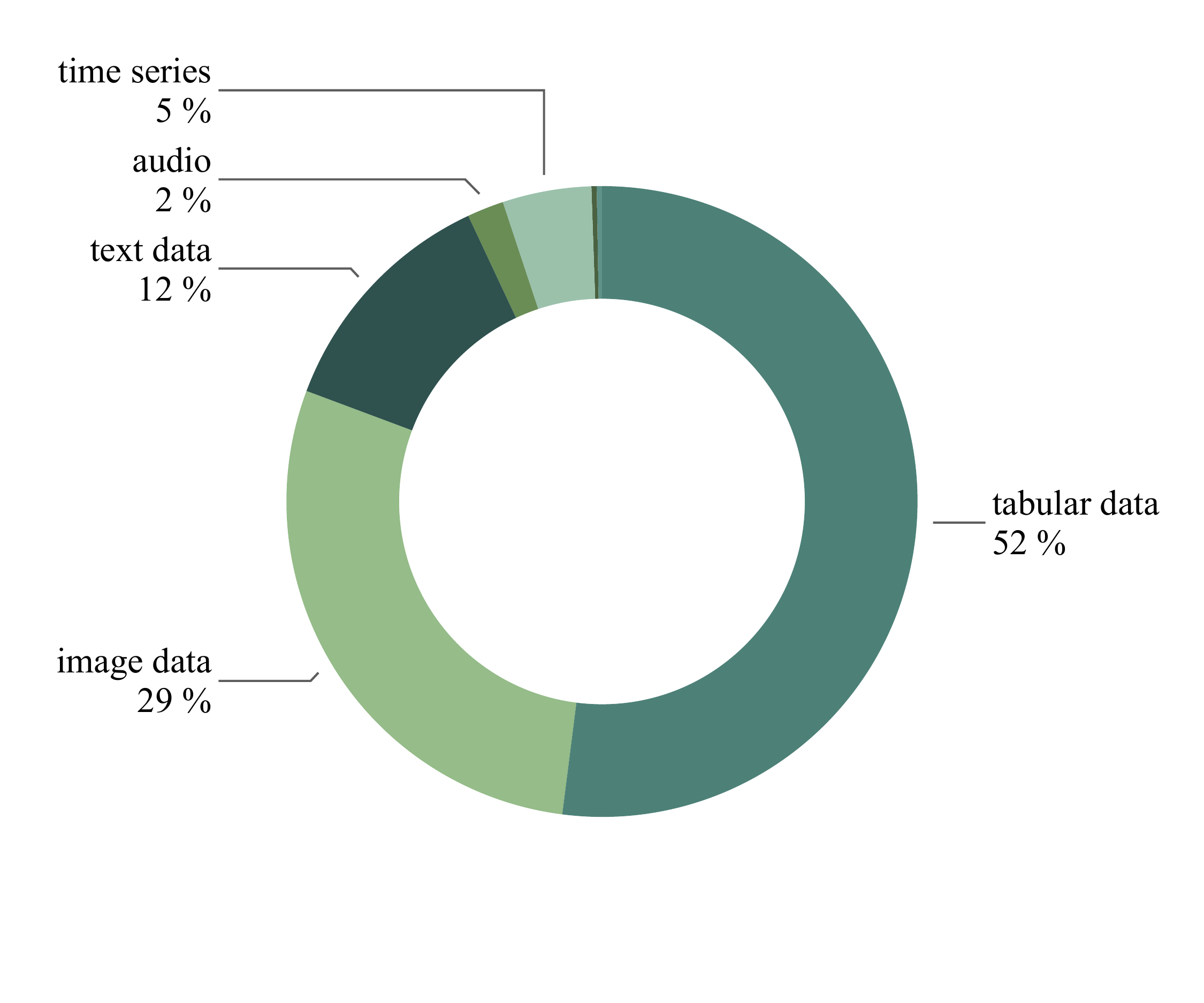}}\\
\caption{The competitions distribution based on the data type.}
\label{fig:data_distr}
  \end{minipage}\hfill
\end{figure}

As a result, our training dataset comprises 395 natural language ML task descriptions paired with 7023 corresponding 
Kaggle solutions. Fig.~\ref{fig:data} overviews the prevalent models featured in the selected solutions. Fig.~\ref{fig:data_distr} illustrates the diversity of data types used in the chosen Kaggle competitions. This work emphasizes ML tasks involving tabular data. However, we do not restrict competitions to numeric tabular formats and consider those involving time series or text data presented with tables.

To assess the effectiveness of our approach, we employ Kaggle competitions that are recent and popular, featuring more than 500 participating teams, ensuring that the tasks were unseen by our model. To approximate the distribution of the training competition space, we randomly select ten machine learning tasks, with a majority operating on numerical data and one each for text, time series, and image data. 

Linguacodus generated instructions validation extends beyond the Kaggle platform, encompassing ML competitions hosted on CodaLab~\citep{CodaLab_competitions_JMLR}. All the data used for validation and testing is not included in the training set.

\subsection*{Baseline}

The overall comparative model for our framework is vanilla GPT-3.5, considering its prominence as a leading tool in natural language generation tasks. While other models exist, such as CodeBERT, CoditT5, PalM-Coder, and CTRL, their suitability for generating code from natural language task descriptions may be limited. Specifically, CodeBERT and CoditT5 are primarily trained for synthesizing code snippets rather than entire pipelines or comprehensive solutions. Therefore, GPT-3.5 is a more relevant and established benchmark in transforming natural language into complete machine learning pipelines. Additionally, GPT-3.5 demonstrates greater efficiency compared to Llama 2~\citep{huggingface} and does not require payment, as GPT-4. Code Llama - Instruct is used as a reference model for the Linguacodus Instruction Creation phase.


\subsection*{Experiments setup and analysis}

In our experiments, we use GPT-3.5 for retrieving instructions from the ML solutions, finding and improving the best instruction, and code generation. The selection of GPT-3.5 is driven by the consideration of balancing quality and inference time using the OpenAI API. However, the framework is generally agnostic to the choice of large language model, allowing for flexibility in utilizing different models based on specific requirements or preferences.

To underscore the significance of the research, we compare the instructions generated by the fine-tuned Llama 2 model and those inferred from Code Llama - Instruct. Our evaluation extends beyond the Kaggle platform, encompassing ML competitions hosted on CodaLab~\citep{CodaLab_competitions_JMLR} to ensure a thorough analysis. All the data used for validation and testing is not included in the training set. We use the selected by Linguacodus best instruction from the top three inferred by Llama 2. Additionally, we include examples of instructions automatically improved with the multi-agent LLM technique through the proposition of more advanced models for training.  

Instructions produced by Code Llama - Instruct generally focus on the high-level approach and conceptual steps involved in training a model. They emphasize data preprocessing, model architecture, and training goals without delving into specific implementation details. In contrast, the fine-tuned Llama 2 instructions provide detailed, step-by-step breakdowns of the data preprocessing, model architecture, and model training processes. While the former offers a broader overview suitable for understanding the overall flow of the task, the latter caters to individuals seeking a more granular understanding, providing a comprehensive guide with specific library references and functions used at each stage (see Appendix~\ref{App B}).

Generating complete and functional code solutions using LLM requires providing the model with a detailed prompt or context outlining the task or problem statement. Hence, well-suited task instructions are vital for code generation. Our pipeline, enhanced by multi-agent LLM, can synthesize code via instructions of predefined quality, making our approach unique and promising for assisting in ML code generation. Appendix~\ref{App C} presents sample code generated by vanilla GPT-3.5 with automatically improved instructions and plain task descriptions. Raw GPT-3.5 output often contains code that cannot be compiled without further specific model training, whereas Linguacodus produces ready-to-run code.

\subsection*{Comparative results}


Table~\ref{tab1} reports the Kaggle scores and percentiles obtained for code generated by Linguacodus and vanilla GPT-3.5 across a selection of randomly chosen machine learning tasks. Table~\ref{tab:id} provides an overview of the mapping between task IDs and corresponding Kaggle competition names. The percentiles reported in Table~\ref{tab1} reflect the relative standing on the Kaggle competition leaderboards, where lower percentiles indicate superior performance. The 0 percentile represents the top ranking, while higher percentiles indicate lower positions on the leaderboard. This comparison provides insight into how the generated solutions perform relative to the broader Kaggle community for each specific competition.

The use of Kaggle leaderboard percentiles provides a comprehensive assessment of the generated models. Unlike traditional code evaluation metrics, such as comparing Abstract Syntax Trees~\citep{knuth1968semantics} or using code similarity measures~\citep{song2024revisiting}, ML task performance requires a more nuanced approach. This is because the goal is to find the most effective solution for a given ML task, which can vary significantly in implementation while achieving similar results. Optimal solutions often emerge from novel combinations of existing ML techniques, making direct code comparison less relevant. Moreover, the effectiveness of generated code can only be truly measured by its performance on the specific ML task.

As shown in Table~\ref{tab1}, Linguacodus consistently produces compilable code, outperforming vanilla GPT-3.5 solutions across specified machine learning metrics. Both Linguacodus and vanilla GPT-3.5 receive natural language descriptions and necessary metadata for each machine learning task as input. To ensure a fair and unbiased comparison, the code generated by both approaches undergoes up to three iterations of error treatment.

\begin{table}[t]
\centering
\caption{The sample results of generated ML code validated on the Kaggle platform. For each competition ID, the comparative scores and percentiles on the Kaggle competition leaderboard are provided. Lower percentiles indicate superior performance, $\times$ denotes an uncompilable solution.}
\begin{tabular}{|c|c|c|c|c|c|c|}
\hline
\makecell{Id}&\makecell{Data \\type}&Metric&\makecell{Linguacodus\\score}&\makecell{Linguacodus  \\percentile}&\makecell{GPT-3.5 \\score}&\makecell{GPT-3.5 \\percentile}\\ \hline
C1&tabular&rmse&\textbf{0.059}&0&$\times$&$\times$\\
C2&tabular&roc-auc&\textbf{0.948}&66&0.943&68\\
C3&\makecell{time \\series}&rmse&\textbf{15.409}&0&24.136&21\\
C4&text&mcrmse&\textbf{0.470}&58&0.548&81\\
C5&tabular&roc-auc&\textbf{0.773}&77&0.752&80\\
C6&image&\makecell{mean \\cosine \\similarity}&\textbf{0.714}&0&$\times$&$\times$\\
C7&tabular&rmse&\textbf{0.578}&69 &0.600&80\\
C8&tabular&mae&\textbf{1.387}&52 &1.978&96\\
C9&tabular&mae&\textbf{366.892}&82 &380.284&93\\
C10&tabular&roc-auc&\textbf{0.862}&76&$\times$&$\times$\\
\hline
\end{tabular}
\label{tab1}
\end{table}

Kaggle, as a competitive platform, traditionally demands significant investment of time and expertise from its participants. Engaging in Kaggle competitions often requires deep understanding of the field and substantial time commitment. Our pipeline for transforming ML task descriptions into code offers a markedly more efficient alternative.

This approach significantly reduces the time and expertise required to bridge the gap between task descriptions and executable code, making machine learning development more accessible. While the OpenAI GPT-3.5 API generates a default solution (without error treatment process) in approximately 6 seconds, our pipeline averages 44 seconds on an A100 GPU. This process involves generating three instructions, correcting them, and sequentially generating code. Despite the longer processing time compared to GPT-3.5, our approach consistently yields superior results. 

\section*{Discussion}
\label{sec:discussions}

As mentioned in 'Related Work', the recent advancements in code generation driven by LLMs have made significant strides, yet several challenges remain. Table~\ref{tab:linguacodus_comparison} discusses how these issues are addressed with Linguacodus.

\begin{table}[h!]
    \centering
    \caption{Comparison of Linguacodus with other Language Models.}
    \begin{tabular}{|p{4cm}|p{11cm}|}
        \hline
        \textbf{Issue} & \textbf{Description} \\
        \hline
        Lack of Specificity & Linguacodus aims to provide more specific and tailored instructions for ML tasks by focusing on high-level information extraction rather than detailed code snippet classification in comparison with LLMs like CodeBERT~\citep{feng2020codebert} and CoditT5 \citep{zhang2022coditt5}. \\
        \hline
        Limited Control Over Code Generation & Ranked instructions allow for a controlled transformation process, providing a structured framework for code generation. While code-related~\citep{anil2023palm}~\citep{rozière2024code} and general-purpose~\citep{openai2023gpt4},~\citep{bubeck2023sparks} LLMs do not offer the generation control tools, Linguacodus' users can choose from the top-ranked instructions, offering control over the generated code. \\
        \hline
        Handling Ambiguity & The ranking process, coupled with the fine-tuning of LLMs, enhances the precision of instructions by prioritizing those that align most closely with the task descriptions, mitigating potential ambiguities, making Linguacodus on par or even better than OpenAI models~\citep{bubeck2023sparks}, but open-source. \\
        \hline
        Inability to Learn Task-Specific Patterns & Leveraging the fine-tuning process with Llama 2 7b on task-specific details allows the model to adapt and learn patterns specific to ML tasks, enhancing the quality and relevance of the generated instructions. \\
        \hline
        Evaluation Metrics and Validation & Compared to evaluation metrics in models such as Texygen~\citep{song2018texygen}, the ranking process, involving evaluation scores and task-specific details, is a robust validation mechanism for the generated instructions, ensuring their alignment with ML tasks and promoting solutions that adhere to evaluation metrics. \\
        \hline
    \end{tabular}
    \label{tab:linguacodus_comparison}
\end{table}



\section*{Limitations}
\label{sec:limitations}
Despite the advancements presented by Linguacodus in addressing the challenges outlined in the section 'Related Work', there are several limitations that warrant consideration. The Code4ML dataset used to train Llama 2, which forms the foundation of Linguacodus, includes competitions only up to 2021. This temporal limitation means that the model may not fully cover the entire range of ML tasks and techniques, particularly recent emergent methods, potentially affecting its performance on cutting-edge problems.

Multi-agent LLM occasionally exhibits suboptimal performance compared to unprocessed Linguacodus instructions, emphasizing the role of context in task's complexity. Ethical considerations surrounding biases and potential misuse of generated code highlight the need for responsible deployment. Linguacodus faces challenges when tasks deviate significantly from those fine-tuned on Llama 2, suggesting a need for dataset enrichment. 

Insufficiently detailed instructions arise when tasks lack comprehensive descriptions, calling for more explicit task information. Recognizing that multi-agent LLM may not consistently outperform initially inferred instructions, human intervention is proposed to select the best instruction. This highlights the need for a balanced approach that combines the strengths of automated models with human judgment in refining outputs.


\section*{Future Work}
\label{sec:future_work}
The temporal limitation of the training dataset underscores the importance of ongoing model updates and the potential for performance gaps in very recent or rapidly evolving areas of machine learning. This observation points to a development of a dynamic framework for enriching the ML data corpus. Such a framework would allow for continuous integration of new ML techniques, datasets, and competition results, ensuring that models like Linguacodus remain current and effective across the evolving landscape of machine learning tasks.

Another promising direction for future work involves exploring alternative, more deterministic approaches to constructing high-level instructions. One such approach is the development of a graph-instruction methodology. This could enable a more structured representation of the ML task, allowing for better assessment of intermediate generation steps and interpretability. By mapping the natural task description to a graph-based representation, we could potentially achieve greater transparency in the instruction generation process, facilitating easier evaluation and refinement of the model's outputs.

\section*{Conclusion}
\label{sec:conclusion}

In this paper, we introduce a comprehensive approach to transforming unstructured ML task descriptions into executable code, presenting the novel Linguacodus model. Leveraging the Code4ML dataset, which encompasses a rich collection of Python code snippets, contest summaries, and data descriptions from Kaggle competitions, our methodology capitalizes on the dataset's valuable competition-related metadata, data types, and scoring metrics. Inspired by the knowledge taxonomy tree introduced in \cite{code4ml}, we adopt a similar organizational framework to achieve dimensional reduction in our ML task description-to-code synthesis approach. However, our approach differs in that it focuses on high-level information extraction rather than individual code snippet classification. This strategic shift simplifies and streamlines the code generation process, making it more efficient and adaptable.\\

Linguacodus is structured into two phases: synthesizing high-level ML solution instructions and transforming these instructions into functional code. 
To generate instructions, the Llama 2 model is fine-tuned on the Code4ML corpus. The top three instructions are then inferred and further refined with the assistance of multi-agent LLM, ensuring the highest quality instructions for subsequent code generation. The second phase involves translating these refined instructions into well-structured and executable code segments, encompassing data preprocessing, model architecture, model training, and submission block generation. This transformation bridges the gap between high-level ML instructions and practical code, ensuring alignment with the underlying ML tasks.\\

Our approach's effectiveness is validated through experiments on Kaggle competitions that are not part of our training data. The results demonstrate that the generated code is compilable and aligns well with the evaluation metrics. We also compare the performance of multi-agent LLM and unprocessed Code Llama - Instructions, highlighting the need for further refinement in multi-agent LLM's algorithmic approach to achieve superior solution quality consistently.\\

In summary, the research provides an innovative and efficient solution for code generation from ML task descriptions, showcasing the capabilities of Linguacodus. By capitalizing on the Code4ML dataset's wealth of resources and introducing a structured approach to instruction synthesis and code generation, we bridge the gap between natural language task descriptions and executable code, making machine learning development more accessible and efficient.

\textbf{CRediT authorship contribution statement}
Ekaterina Trofimova: Conceptualization, Investigation, Methodology, Software, Validation, Formal analysis, Writing - Original Draft, Writing - Review \& Editing, Visualization, Supervision, Project administration. Emil Sataev: Investigation, Software, Methodology, Data Curation, Formal analysis. Andrey E. Ustyuzhanin: Conceptualization, Supervision, Formal analysis, Funding acquisition, Methodology, Writing – Review \& Editing.





\textbf{Acknowledgments}
We would like to express our appreciation to Denis Derkach and Artem Maevskiy for their invaluable comments and support.


\bibliographystyle{elsarticle-harv}



\begin{thebibliography}{00}

\bibitem[Achiam et al., 2023]{openai2023gpt4}
Achiam, J., Adler, S., Agarwal, S., Ahmad, L., Akkaya, I., Aleman, F. L., , ...  McGrew, B., (2023). Gpt-4 technical report. arXiv preprint arXiv:2303.08774.

\bibitem[Agashe et al., 2019]{agashe2019juice}
Agashe, R., Iyer, S., Zettlemoyer, L. (2019). JuICe: A large scale distantly supervised dataset for open domain context-based code generation. arXiv preprint arXiv:1910.02216.


\bibitem[Allamanis et al., 2018]{allamanis2018survey} {Allamanis, M., Barr, E. T., Devanbu, P., Sutton, C. (2018). A survey of machine learning for big code and naturalness. ACM Computing Surveys (CSUR), 51(4), 1-37.}

\bibitem[Alpaydin, 2021]{alpaydin2021machine}
Alpaydin, E. (2021). Machine learning. MIT press.

\bibitem[Anil et al., 2023]{anil2023palm}
{Anil, R., Dai, A. M., Firat, O., Johnson, M., Lepikhin, D., Passos, A., ... \& Wu, Y. (2023). Palm 2 technical report. arXiv preprint arXiv:2305.10403.}

\bibitem[Buse and Weimer, 2012]{buse2012synthesizing}
Buse, R. P.,  Weimer, W. (2012, June). Synthesizing API usage examples. In 2012 34th International Conference on Software Engineering (ICSE) (pp. 782-792). IEEE.

\bibitem[Berezovskiy et al., 2023]{code4mlclass}
{Berezovskiy, V., Gorodilova, A., Trofimova, E., \& Ustyuzhanin, A. (2023). Machine learning code snippets semantic classification. PeerJ Computer Science, 9, e1654.}

\bibitem[Bubeck et al., 2023]{bubeck2023sparks}
{Bubeck, S., Chandrasekaran, V., Eldan, R., Gehrke, J., Horvitz, E., Kamar, E., ... \& Zhang, Y. (2023). Sparks of artificial general intelligence: Early experiments with gpt-4. arXiv preprint arXiv:2303.12712.}

\bibitem[Chen et al., 2021]{Chen2021EvaluatingLL}
{Chen, M., Tworek, J., Jun, H., Yuan, Q., Pinto, H. P. D. O., Kaplan, J., ... \& Zaremba, W. (2021). Evaluating large language models trained on code. arXiv preprint arXiv:2107.03374.}

\bibitem[Chandel et al., 2022]{chandel2022training}
{Chandel, Sh., Clement, C. B, Serrato, G., Sundaresan, N. 2022. Training and evaluating a jupyter notebook data science assistant. arXiv preprint arXiv:2201.12901.}

\bibitem[Chowdhery et al., 2023]{chowdhery2023palm}
{Chowdhery, A., Narang, S., Devlin, J., Bosma, M., Mishra, G., Roberts, A., ... \& Fiedel, N. (2023). Palm: Scaling language modeling with pathways. Journal of Machine Learning Research, 24(240), 1-113.}

\bibitem[Desai et al., 2016]{desai2016program} Desai, A., Gulwani, S., Hingorani, V., Jain, N., Karkare, A., Marron, M., Roy, S. (2016, May). Program synthesis using natural language. In Proceedings of the 38th International Conference on Software Engineering (pp. 345-356).

\bibitem[Devlin et al., 2018]{devlin2018bert}
Devlin, J., Chang, M. W., Lee, K., Toutanova, K. (2018). Bert: Pre-training of deep bidirectional transformers for language understanding. arXiv preprint arXiv:1810.04805.

\bibitem[Drozdova et al., 2023]{code4ml}
{Drozdova, A., Trofimova, E., Guseva, P., Scherbakova, A., \& Ustyuzhanin, A. (2023). Code4ML: a large-scale dataset of annotated Machine Learning code. PeerJ Computer Science, 9, e1230.}

\bibitem[Erickson et al., 2020]{erickson2020autogluontabular}
{Erickson, N., Mueller, J., Shirkov, A., Zhang, H., Larroy, P., Li, M., \& Smola, A. (2020). Autogluon-tabular: Robust and accurate automl for structured data. arXiv preprint arXiv:2003.06505.}

\bibitem[Feng et al., 2020]{feng2020codebert}
{Feng, Z., Guo, D., Tang, D., Duan, N., Feng, X., Gong, M., ... \& Zhou, M. (2020). Codebert: A pre-trained model for programming and natural languages. arXiv preprint arXiv:2002.08155.}

\bibitem[Franklin, 2023]{commonlit-evaluate-student-summaries}
Franklin, A. (2023). CommonLit - Evaluate Student Summaries. Kaggle. Retrieved from https://kaggle.com/competitions/commonlit-evaluate-student-summaries

\bibitem[Gulwani et al., 2023]{gulwani2017program} Gulwani, S., Polozov, O., Singh, R. (2017). Program synthesis. Foundations and Trends® in Programming Languages, 4(1-2), 1-119.

\bibitem[Zheng et al., 2024]{huggingface}
Zheng, L., Chiang, W. L., Sheng, Y., Zhuang, S., Wu, Z., Zhuang, Y., ... \& Stoica, I. (2024). Judging llm-as-a-judge with mt-bench and chatbot arena. Advances in Neural Information Processing Systems, 36.

\bibitem[Harel et al., 1990]{harel1990statemate}
Harel, D., Lachover, H., Naamad, A., Pnueli, A., Politi, M., Sherman, R., ... \& Trakhtenbrot, M. (1990). Statemate: A working environment for the development of complex reactive systems. IEEE Transactions on software engineering, 16(4), 403-414.

\bibitem[Jung, 2022]{jung2022machine}
Jung, A. (2022). Machine learning: the basics. Springer Nature.

\bibitem[Keskar et al., 2019]{keskar2019ctrl}
{Keskar, N. S., McCann, B., Varshney, L. R., Xiong, C., \& Socher, R. (2019). Ctrl: A conditional transformer language model for controllable generation. arXiv preprint arXiv:1909.05858.}

\bibitem[Kluyver et al., 2016]{kluyver2016jupyter}
Kluyver, T., Ragan-Kelley, B., Pérez, F., Granger, B. E., Bussonnier, M., Frederic, J., ... \& Willing, C. (2016). Jupyter Notebooks-a publishing format for reproducible computational workflows. Elpub, 2016, 87-90.

\bibitem[Knuth, 1968]{knuth1968semantics} {Knuth, D. E. (1968). Semantics of context-free languages. Mathematical systems theory,  2(2), 127–145}

\bibitem[Lei et al., 2013]{lei2013natural}
Lei, T., Long, F., Barzilay, R., Rinard, M. (2013, August). From natural language specifications to program input parsers. In Proceedings of the 51st Annual Meeting of the Association for Computational Linguistics (Volume 1: Long Papers) (pp. 1294-1303).

\bibitem[LeDell and Poirier, 2020]{LeDell2020H2OAS}
{LeDell, E., \& Poirier, S. (2020, July). H2o automl: Scalable automatic machine learning. In Proceedings of the AutoML Workshop at ICML (Vol. 2020). San Diego, CA, USA: ICML.}

\bibitem[Liu et al., 2020]{liu2020deep}
Liu, H., Shen, M., Zhu, J., Niu, N., Li, G., Zhang, L. (2020). Deep learning based program generation from requirements text: Are we there yet?. IEEE Transactions on Software Engineering, 48(4), 1268-1289.

\bibitem[Ling et al., 2016]{ling2016latent}
Ling, W., Grefenstette, E., Hermann, K. M., Kočiský, T., Senior, A., Wang, F., Blunsom, P. (2016). Latent predictor networks for code generation. arXiv preprint arXiv:1603.06744.

\bibitem[Li et al., 2022]{doi:10.1126/science.abq1158}
{Li, Y., Choi, D., Chung, J., Kushman, N., Schrittwieser, J., Leblond, R., ... \& Vinyals, O. (2022). Competition-level code generation with alphacode. Science, 378(6624), 1092-1097.}

\bibitem[Moruri et al., 2023]{playground-series-s3e20}
{Moruri, D., Bray, A., Reade, W., \& Chow, A. (2023). Predict CO2 Emissions in Rwanda. Kaggle. Retrieved from https://kaggle.com/competitions/playground-series-s3e20}

\bibitem[Nair et al., 2023]{nair2023dera}
{Nair, V., Schumacher, E., Tso, G., \& Kannan, A. (2023). DERA: enhancing large language model completions with dialog-enabled resolving agents. arXiv preprint arXiv:2303.17071.}

\bibitem[Pavao et al., 2023]{CodaLab_competitions_JMLR} {Pavao, A., Guyon, I., Letournel, A. C., Tran, D. T., Baro, X., Escalante, H. J., ... \& Xu, Z. (2023). Codalab competitions: An open source platform to organize scientific challenges. Journal of Machine Learning Research, 24(198), 1-6.}

\bibitem[Raychev et al., 2014]{raychev2014code}
Raychev, V., Vechev, M., Yahav, E. (2014, June). Code completion with statistical language models. In Proceedings of the 35th ACM SIGPLAN conference on programming language design and implementation (pp. 419-428).

\bibitem[Rabinovich et al., 2017]{rabinovich2017abstract}
Rabinovich, M., Stern, M., Klein, D. (2017). Abstract syntax networks for code generation and semantic parsing. arXiv preprint arXiv:1704.07535.

\bibitem[Radford et al., 2018]{radford2018improving}
Radford, A., Narasimhan, K., Salimans, T., Sutskever, I. (2018). Improving language understanding by generative pre-training.

\bibitem[Reade and Chow, 2023a]{playground-series-s3e15}
{Reade, W., \& Chow, A. (2023a). Feature Imputation with a Heat Flux Dataset. Kaggle. Retrieved from https://kaggle.com/competitions/playground-series-s3e15}

\bibitem[Reade and Chow, 2023b]{playground-series-s3e17}
{Reade, W., \& Chow, A. (2023b). Binary Classification of Machine Failures. Kaggle. Retrieved from https://kaggle.com/competitions/playground-series-s3e17}

\bibitem[Reade and Chow, 2023c]{playground-series-s3e4}
{Reade, W., \& Chow, A. (2023c). Binary Classification with a Tabular Credit Card Fraud Dataset. Kaggle. Retrieved from https://kaggle.com/competitions/playground-series-s3e4}

\bibitem[Reade and Chow, 2023d]{playground-series-s3e1}
{Reade, W., \& Chow, A. (2023d). Regression with a Tabular California Housing Dataset. Kaggle. Retrieved from https://kaggle.com/competitions/playground-series-s3e1}

\bibitem[Reade and Chow, 2023e]{playground-series-s3e16}
{Reade, W., \& Chow, A. (2023e). Regression with a Crab Age Dataset. Kaggle. Retrieved from https://kaggle.com/competitions/playground-series-s3e16}

\bibitem[Reade and Chow, 2023f]{playground-series-s3e6}
{Reade, W., \& Chow, A. (2023f). Regression with a Tabular Paris Housing Price Dataset. Kaggle. Retrieved from https://kaggle.com/competitions/playground-series-s3e6}

\bibitem[Reade and Chow, 2023g]{playground-series-s3e14}
{Reade, W., \& Chow, A. (2023g). Regression with a Wild Blueberry Yield Dataset. Kaggle. Retrieved from https://kaggle.com/competitions/playground-series-s3e14}

\bibitem[Reade and Chow, 2024]{playground-series-s4e1}
{Reade, W., \& Chow, A. (2024). Binary Classification with a Bank Churn Dataset. Kaggle. Retrieved from https://kaggle.com/competitions/playground-series-s4e1}

\bibitem[Reade et al., 2023]{stable-diffusion-image-to-prompts}
{Reade, W., Cukierski, W., \& Chow, A. (2023). "Stable Diffusion - Image to Prompts." Kaggle. Retrieved from https://kaggle.com/competitions/stable-diffusion-image-to-prompts}

\bibitem[Rozière et al., 2024]{rozière2024code}
{Roziere, B., Gehring, J., Gloeckle, F., Sootla, S., Gat, I., Tan, X. E., ... \& Synnaeve, G. (2023). Code llama: Open foundation models for code. arXiv preprint arXiv:2308.12950.}


\bibitem[Shen et al., 2024]{shen2023hugginggpt}
{Shen, Y., Song, K., Tan, X., Li, D., Lu, W., \& Zhuang, Y. (2024). Hugginggpt: Solving ai tasks with chatgpt and its friends in hugging face. Advances in Neural Information Processing Systems, 36.}

\bibitem[Song et al., 2024]{song2024revisiting}
{Song, Y., Lothritz, C., Tang, D., Bissyand{\'e}, T., Klein, J. (2024). Revisiting Code Similarity Evaluation with Abstract Syntax Tree Edit Distance. arXiv preprint arXiv:2404.08817}

\bibitem[Touvron et al., 2023]{touvron2023llama}
{Touvron, H., Martin, L., Stone, K., Albert, P., Almahairi, A., Babaei, Y., ... \& Scialom, T. (2023). Llama 2: Open foundation and fine-tuned chat models. arXiv preprint arXiv:2307.09288.}

\bibitem[Trofimova et al., 2024]{trofimova2024linguacodus}
{Trofimova, E., Sataev, E., Ustyuzhanin, A. E. (2024). Linguacodus: A Synergistic Framework for Transformative Code Generation in Machine Learning Pipelines. arXiv preprint arXiv:2403.11585}

\bibitem[Vaswani A. et al., 2017]{vaswani2017attention}
{Vaswani, A., Shazeer, N., Parmar, N., Uszkoreit, J., Jones, L., Gomez, A. N., Kaiser, {\L}, Polosukhin, I., 2017. Attention is all you need. Advances in neural information processing systems, 30.}

\bibitem[Vakhrushev et al., 2021]{vakhrushev2022lightautoml}
{Vakhrushev, A., Ryzhkov, A., Savchenko, M., Simakov, D., Damdinov, R., \& Tuzhilin, A. (2021). Lightautoml: Automl solution for a large financial services ecosystem. arXiv preprint arXiv:2109.01528.}

\bibitem[Vaithilingam et al., 2021]{vaithilingam2022expectation}{Vaithilingam, P., Zhang, T., \& Glassman, E. L. (2022, April). Expectation vs. experience: Evaluating the usability of code generation tools powered by large language models. In Chi conference on human factors in computing systems extended abstracts (pp. 1-7).}

\bibitem[Winograd., 1972]{winograd1972understanding}
Winograd, T. (1972). Understanding natural language. Cognitive psychology, 3(1), 1-191.

\bibitem[Wang Sh. et al., 2023]{wang2023natural}
{Wang, S., Geng, M., Lin, B., Sun, Z., Wen, M., Liu, Y., ... \& Mao, X. (2023, November). Natural Language to Code: How Far Are We?. In Proceedings of the 31st ACM Joint European Software Engineering Conference and Symposium on the Foundations of Software Engineering (pp. 375-387).}

\bibitem[Wen. et al., 2024]{wen2024grounding}
Wen, Y., Yin, P., Shi, K., Michalewski, H., Chaudhuri, S., Polozov, A. (2024). Grounding Data Science Code Generation with Input-Output Specifications. arXiv preprint arXiv:2402.08073.

\bibitem[Yin and Neubig, 2017]{yin2017syntactic}
Yin, P., Neubig, G. (2017). A syntactic neural model for general-purpose code generation. arXiv preprint arXiv:1704.01696.

\bibitem[Yin and Neubig, 2018]{yin2018tranx}
Yin, P., Neubig, G. (2018). TRANX: A transition-based neural abstract syntax parser for semantic parsing and code generation. arXiv preprint arXiv:1810.02720.

\bibitem[Yin et al., 2022]{yin2022natural}
Yin, P., Li, W. D., Xiao, K., Rao, A., Wen, Y., Shi, K., ... \& Sutton, C. (2022). Natural language to code generation in interactive data science notebooks. arXiv preprint arXiv:2212.09248.

\bibitem[Zhang et al., 2022]{zhang2022coditt5}
{Zhang, J., Panthaplackel, S., Nie, P., Li, J. J., \& Gligoric, M. (2022, October). Coditt5: Pretraining for source code and natural language editing. In Proceedings of the 37th IEEE/ACM International Conference on Automated Software Engineering (pp. 1-12).}

\bibitem[Zhu et al., 2018]{song2018texygen}
{Zhu, Y., Lu, S., Zheng, L., Guo, J., Zhang, W., Wang, J., \& Yu, Y. (2018, June). Texygen: A benchmarking platform for text generation models. In The 41st international ACM SIGIR conference on research \& development in information retrieval (pp. 1097-1100).}

\end{thebibliography}

\appendix

\renewcommand\thesection{\Alph{section}}
\renewcommand\thelstlisting{\thesection.\arabic{lstlisting}}

\newcommand{\sectionbreak}{\setcounter{table}{0}} 
\renewcommand\thetable{\Alph{section}.\arabic{table}}
\renewcommand\thefigure{\Alph{section}.\arabic{figure}}

\section{Llama 2 fine-tuning details}
\label{App A}

To align natural language descriptions of machine learning tasks with high-level code instructions extracted from ML code solutions, we fine-tune the Llama 2 model. Table~\ref{table} presents the hyperparameters used in the Llama 2 fine-tuning process.

\begin{table}
\caption{Llama 2 fine-tuning hyper-parameters.}
\label{table}
\centering
\begin{tabular}{|l|c|}
\hline
\multicolumn{2}{|c|}{\textbf{LoRA Parameters}} \\
\hline
LoRA attention dimension & 64 \\
Alpha parameter for LoRA scaling & 16 \\
Dropout probability for LoRA layers & 0.1 \\
\hline
\multicolumn{2}{|c|}{\textbf{4-Bit Precision Parameters}} \\
\hline
4-bit precision base model loading & True \\
Compute dtype for 4-bit base models & float16 \\
Quantization type & nf4 \\
Nested quantization for 4-bit base models & False \\
\hline
\multicolumn{2}{|c|}{\textbf{TrainingArguments Parameters}} \\
\hline
Number of training epochs & 1 \\
Enable fp16/bf16 training & False/False \\
Batch size per GPU for training & 4 \\
Batch size per GPU for evaluation & 4 \\
Enable gradient checkpointing & True \\
Maximum gradient normal & 0.3 \\
Initial learning rate & 2e-4 \\
Weight decay & 0.001 \\
Optimizer & AdamW \\
Learning rate schedule & constant \\
Number of training steps & -1 \\
Ratio of steps for a linear warmup & 0.03 \\
Group sequences into same length batches & True \\
Save checkpoint every X updates steps & 500 \\
Log every X updates steps & 25 \\
\hline
\multicolumn{2}{|c|}{\textbf{Sequence Fine-Tuning Parameters}} \\
\hline
Maximum sequence length & None \\
\hline
\end{tabular}
\end{table}

\section{Sample instructions inferred by Code Llama - Instruct and fine-tuned Llama 2}
\label{App B}

This section presents a comparative analysis of instructions for various machine learning tasks generated by three methods: Code Llama - Instruct; fine-tuned Llama 2 (best instructions selected by Linguacodus); multi-agent LLM automatic improvement.

Our analysis focuses on four competitions: two from CodaLab and two from Kaggle. Table~\ref{table_names} summarizes the key information for these selected competitions. The set of tasks represented in these competitions allows for a comprehensive comparison. We use the competition names and task descriptions as prompts for instruction generation.

\begin{table*}[t]
\centering
\caption{Competitions information.}
\label{table_names}
\centering
\setlength{\tabcolsep}{5pt}
\begin{tabular}{|p{85pt}|p{40pt}|p{40pt}|p{40pt}|p{180pt}|}
\hline
Competition name &Source & Data Type & Metric Type& Task description\\
\hline
"SHROOM - a Shared-task on Hallucinations and Related Observable Overgeneration Mistake"~\citep{CodaLab_competitions_JMLR} &CodaLab & text & accuracy& SHROOM challenges participants to identify outputs that are grammatically correct but semantically inaccurate, in a binary classification task. The competition focuses on detecting "hallucinations" in outputs from neural language models across definition modeling, machine translation, and paraphrase generation, with or without access to the model that generated these outputs\\
\hline
"Climate Activism Stance and Hate Event Detection Shared Task"~\citep{CodaLab_competitions_JMLR}&CodaLab & text & f1 score&The CASE 2024 competition at EACL 2024 focuses on three subtasks in the context of climate activism: identifying hate speech, detecting targets of hate speech (individuals, organizations, communities), and determining the stance (support, oppose, neutral) of given texts.\\
\hline
"Stable Diffusion - Image to Prompts"~\citep{stable-diffusion-image-to-prompts}&Kaggle & images & mean cosine similarity&The competition aims to develop a model that can predict the text prompt that corresponds to a given generated image, challenging the understanding of the relationship between text prompts and images in text-to-image models. The goal is to create embeddings for predicted prompts, ensuring robust prompt similarity assessment, and exploring prompt engineering in the context of image generation.\\
\hline
"Regression with a Tabular Paris Housing Price Dataset"~\citep{playground-series-s3e6}&Kaggle & tabular & rmse& Regression with a Tabular Paris Housing Price Dataset\\
\hline
\end{tabular}
\end{table*}

Tables~\ref{tab:CodaLab0},~\ref{tab:CodaLab0_our},~\ref{tab:CodaLab0_smart} showcase the retrieved instructions for CodaLab competition "SHROOM - a Shared-task on Hallucinations and Related Observable Overgeneration Mistake". The first instruction adopts a high-level approach, focusing on the overarching strategy and conceptual steps involved in training a model to identify "hallucinations" in neural language model outputs. It significantly emphasizes data preprocessing, model architecture, and training objectives while avoiding intricate implementation details.

In contrast, the subsequent instruction provides a meticulous, step-by-step breakdown of the data preprocessing, model architecture, and model training procedures. It intricately outlines actions such as data loading using pandas, tokenization with the specific Tokenizer class, fine-tuning a BERT-based model through the TensorFlow BERT library, and tracking training progress using the TensorFlow TensorBoard API. This level of detail makes it more suitable for direct implementation.

An instruction automatically improved with multi-agent LLM goes beyond by suggesting a more advanced model for solving an ML task. Additionally, it specifies the optimization algorithm and training metric, providing a more refined and advanced set of instructions for users. Thus, the potential of multi-agent LLM in enhancing the quality and sophistication of instructions for machine learning tasks.

\begin{table}[ht]
\centering
\caption{A sample instruction generated for "SHROOM - a Shared-task on Hallucinations and Related Observable Overgeneration Mistake" competition by Code Llama - Instruct.}
\begin{tabular}{p{0.95\linewidth}}
\hline
As a data analyst, I would start by preprocessing the text data to
ensure that it is in a format that can be easily analyzed and processed.
This may involve removing stop words, stemming or lemmatizing the
text, and converting the text to a numerical representation using
techniques such as bag-of-words or word embeddings.\\
\hline
Once the data is preprocessed, I would then build a binary
classification model that can accurately detect grammatically correct
but semantically inaccurate outputs. This model architecture could
include a combination of techniques such as word embeddings,
recurrent neural networks (RNNs), and convolutional neural networks
(CNNs) to capture the nuances of language and detect patterns in the
data.\\
\hline
To train the model, I would use a dataset of labeled examples, where
each example is annotated with a binary label indicating whether the
output is grammatically correct but semantically inaccurate or not. I
would then use this dataset to train the model using a suitable machine
learning algorithm, such as logistic regression or support vector
machines (SVMs), and evaluate its performance using metrics such as
accuracy, precision, recall, and F1 score.\\
\hline
Overall, the goal of this task is to develop a model that can accurately
detect "hallucinations" in outputs from neural language models, and to
evaluate its performance using appropriate metrics. By preprocessing
the data and using a suitable model architecture, I believe that I can
develop a model that can accurately detect these hallucinations and
improve the quality of the outputs generated by the neural language
models.\\
\hline
\end{tabular}
\label{tab:CodaLab0}
\end{table}

\begin{table}[ht]
\centering
\caption{The best out of top-3 sample instruction generated for "SHROOM - a Shared-task on Hallucinations and Related Observable Overgeneration Mistake" competition by fine-tuned Llama 2.}
\begin{tabular}{p{0.95\linewidth}}
\hline
Data Preprocessing:\\
- The training data is loaded from a CSV file using pandas.\\
- The test data is loaded from a CSV file using pandas.\\
- The training data is split into training and validation sets using the train\_test\_split function from sklearn.\\
- The training data is further processed by converting the text data into tokenized sequences using the Tokenizer class from nltk.\\
- The tokenized sequences are then converted into input IDs and attention masks using the TensorFlow SequenceClassification API.\\
\hline
Model Architecture:\\
- The model architecture used is a BERT-based model.\\
- The BERT model is loaded from a pre-trained checkpoint using the BERTCheckpoint class from the TensorFlow BERT library.\\
- The BERT model is then fine-tuned on the training data using the TensorFlow TensorBoard API.\\
- The fine-tuned BERT model is used as the base model for the classification task.\\
\hline
Model Training:\\
- The model is trained using the TensorFlow TensorBoard API.\\
- The training data is used to train the model for a specified number of epochs.\\
- The training progress is monitored using the TensorFlow TensorBoard API.\\
- The trained model is then used to make predictions on the test data.\\
- The predictions are saved in a submission file for evaluation.\\
\hline
\end{tabular}
\label{tab:CodaLab0_our}
\end{table}

\begin{table}[ht]
\centering
\caption{An automatically chosen and improved with multi-agent LLM best sample instruction generated for "SHROOM - a Shared-task on Hallucinations and Related Observable Overgeneration Mistake" competition by fine-tuned Llama 2.}
\begin{tabular}{p{0.95\linewidth}}
\hline
Data Preprocessing:\\
• Load the training and test data from CSV files, ensuring that the data is properly formatted. \\
• Preprocess the text data to prepare it for model input. This may include tokenization and data cleaning if necessary.\\
\hline
Model Architecture:\\
• Choose an appropriate model architecture for this binary classification task. Consider using pre-trained models like BERT or GPT-3 as a starting point.\\
• Fine-tune the selected model on the training data to adapt it to the specific task. Ensure that the architecture includes appropriate layers for text processing and classification.\\
\hline
Model Training:\\
• Set up the training process with the following components:\\
◦  Loss Function: Use binary cross-entropy as the loss function for binary classification. \\
◦  Optimizer: Select an optimizer like Adam.\\
◦  Metrics: Track accuracy as the evaluation metric.\\
• Train the model for a specified number of epochs, monitoring the training progress to avoid overfitting.\\
• Split the data into training and validation sets to evaluate the model's performance during training.\\
\hline
Model Evaluation:\\
• After training, evaluate the model on the validation set to assess its accuracy and other relevant metrics.\\
• Make necessary adjustments to the model or training process based on the evaluation results.\\
\hline
Model Evaluation:\\
• Use the trained and validated model to make predictions on the test data. \\
• Save the predictions in a submission file for evaluation.\\
\hline
\end{tabular}
\label{tab:CodaLab0_smart}
\end{table}

Tables~\ref{tab:CodaLab2}--~\ref{tab:kaggle3_smart} depict the triple instructions for the competitions "Climate Activism Stance and Hate Event Detection Shared Task", "Stable Diffusion - Image to Prompts" and "Regression with a Tabular Paris Housing Price Dataset", respectively. Once again, the fine-tuned Llama 2 yields precise and straightforward instructions for generating accurate code. Notably, these instructions are the optimal choice for the assigned tasks, aligning seamlessly with the prompt specifications and achieving the highest solution rating.

multi-agent LLM enriches the instructions by delving deeper and justifying the ML steps. This enhancement contributes to the clarity of the instructions and adds a layer of transparency, aiding users in understanding the rationale behind the suggested approach. The collaborative synergy between fine-tuned Llama 2 and multi-agent LLM demonstrates the potential for leveraging advanced models to refine and augment machine-generated instructions.

\begin{table}[ht]
\centering
\caption{A sample instruction generated for "Climate Activism Stance and Hate Event Detection Shared Task" competition by Code Llama - Instruct.}
\begin{tabular}{p{0.95\linewidth}}
\hline
As a data analyst, I would start by preprocessing the data to ensure that it is in a format that can be easily analyzed and processed. This may involve cleaning the data, removing any irrelevant or redundant information, and transforming the data into a format that can be easily analyzed by a machine learning model.\\
\hline
Once the data is preprocessed, I would then move on to designing the model architecture. For the task of identifying hate speech, detecting targets of hate speech, and determining the stance of given texts, a combination of natural language processing (NLP) and machine learning techniques would be necessary.\\
\hline
To identify hate speech, I would use a technique called sentiment analysis, which involves analyzing the sentiment of a piece of text to determine whether it is positive, negative, or neutral. To detect targets of hate speech, I would use a technique called entity recognition, which involves identifying and extracting specific entities such as individuals, organizations, and communities from the text.\\ \\
To determine the stance of given texts, I would use a technique called stance detection, which involves analyzing the sentiment of a piece of text to determine whether it is supportive, opposing, or neutral.\\
\hline
Once the model architecture is designed, I would then train the model using the preprocessed data. The training process would involve feeding the preprocessed data into the model and adjusting the model's parameters to minimize the error between the model's predictions and the true labels.\\ \\
Throughout the training process, I would monitor the model's performance using metrics such as accuracy, precision, recall, and F1-score. Once the model is trained, I would evaluate its performance on a separate test set to ensure that it is generalizing well to new data.\\
\hline
Overall, the key to success in this machine learning task is to carefully preprocess the data, design an appropriate model architecture, and train the model using a large and diverse dataset. By following these steps, I believe that it is possible to achieve high performance on the CASE 2024 competition at EACL 2024.\\
\hline
\end{tabular}
\label{tab:CodaLab2}
\end{table}

\begin{table}[ht]
\centering
\caption{A sample instruction generated for "Climate Activism Stance and Hate Event Detection Shared Task" competition by Code Llama - Instruct.}
\begin{tabular}{p{0.95\linewidth}}
\hline
Data Preprocessing:\\
- Read the training and test data from CSV files.\\
- Perform data cleaning and preprocessing steps, such as handling missing values, converting categorical variables to numerical codes, and encoding text data.\\
- Split the data into training and testing sets.\\
\hline
Model Architecture:\\
- Define the model architecture using a pre-trained language model (e.g., BERT) and a classification head.\\
- The model will be trained using a combination of binary cross-entropy loss and categorical cross-entropy loss.\\
\hline
Model Training:\\
- Use the training data to train the model.\\
- Use the validation data to evaluate the model's performance.\\ 
- Use the testing data to make predictions\\
\hline
\end{tabular}
\label{tab:CodaLab2_our}
\end{table}

\begin{table}[ht]
\centering
\caption{An automatically chosen and improved with multi-agent LLM best sample instruction generated for "Climate Activism Stance and Hate Event Detection Shared Task" competition by fine-tuned Llama 2.}
\begin{tabular}{p{0.95\linewidth}}
\hline
Data Preprocessing:\\
• Clean and normalize the data, removing irrelevant or redundant information. \\
• Handle missing values and outliers appropriately.\\
\hline
Model Architecture:\\
• Select a suitable machine learning algorithm for text classification based on the task's requirements and data characteristics. Consider using natural language processing (NLP) techniques.\\
\hline
Data Splitting:\\
• Split the dataset into training, validation, and test sets to ensure proper model evaluation.\\
\hline
Model Training and Hyperparameter Tuning:\\
• Train the model on the training dataset, adjusting parameters to minimize error. Perform hyperparameter tuning to optimize model performance.\\
\hline
Evaluation:\\
• Calculate the F1-score to evaluate the model's performance, ensuring it aligns with the competition's metric type\\
\hline
Handling Imbalanced Data:\\
• Implement techniques to address imbalanced class distribution if necessary.\\
\hline
\end{tabular}
\label{tab:CodaLab2_smart}
\end{table}

\begin{table}[ht]
\centering
\caption{A sample instruction generated for "Stable Diffusion - Image to Prompts" competition by Code Llama - Instruct.}
\begin{tabular}{p{0.95\linewidth}}
\hline
As a data analyst, I would start by preprocessing the image data to ensure that it is in a format that can be used for machine learning. This may involve resizing the images, normalizing the pixel values, and possibly applying data augmentation techniques to increase the size of the dataset. \\ 
Once the data is preprocessed, I would then design a model architecture that is suitable for the task at hand. This may involve using a combination of convolutional neural networks (CNNs) and recurrent neural networks (RNNs) to capture the relationships between text prompts and images.\\
\hline
To train the model, I would use a mean cosine similarity metric to measure the similarity between the predicted prompts and the actual prompts. I would also use a technique called prompt engineering to ensure that the predicted prompts are robust and accurate. This involves using a combination of techniques such as word embedding, attention mechanisms, and recurrent neural networks to improve the performance of the model.\\
\hline
Overall, the goal of this task is to develop a model that can predict the text prompt that corresponds to a given generated image, and to explore the relationship between text prompts and images in the context of text-to-image models. By using a combination of data preprocessing, model architecture, and model training, I believe that we can achieve this goal and make significant contributions to the field of natural language processing and computer vision.\\
\hline
\end{tabular}
\label{tab:kaggle1}
\end{table}

\begin{table}[ht]
\centering
\caption{The best out of top-3 sample instruction generated for "Stable Diffusion - Image to Prompts" competition by fine-tuned Llama 2.}
\begin{tabular}{p{0.95\linewidth}}
\hline
Data Preprocessing:\\
• The training data consists of 100,000 images and their corresponding text prompts. The test data consists of 10,000 images and their corresponding text prompts.\\
• The images are resized to a fixed size of 224x224 pixels.
• The text prompts are tokenized using the tokenizer from the BERT model.\\
• The tokenized text prompts are converted to integers using the tokenizer's $convert\_tokens\_to\_ids$ method.\\
• The image IDs and text prompt IDs are stored in separate arrays.\\
\hline
Model Architecture:\\
• The model architecture used is a BERT-based model.\\
• The BERT model is loaded from a pre-trained checkpoint.\\
• The BERT model is used as a feature extractor.\\
• The extracted features are passed through a linear layer with a sigmoid activation function. \\
• The sigmoid activation function is used to convert the output of the linear layer to a probability value.\\
\hline
Model Training:\\
• The model is trained using the training data.\\
• The training data is split into training and validation sets.\\
• The model is trained using the Adam optimizer and a learning rate of 1e-5. The training process is repeated for a specified number of epochs.\\
• The model is evaluated on the validation set during training.\\
• The training and validation loss and accuracy are recorded.\\
\hline
Additional Information:\\
• The model is saved after training.\\
• The saved model is loaded for use in the test data.\\
• The test data is passed through the model to generate predictions.\\
• The predicted text prompts are stored in a list.\\
• The predicted text prompts are then converted to integers using the tokenizer's $convert\_tokens\_to\_ids$ method.\\
• The predicted text prompts are saved in a submission file.\\
\hline
\end{tabular}
\label{tab:kaggle1_our}
\end{table}

\begin{table}[ht]
\centering
\caption{An automatically chosen and improved with multi-agent LLM best sample instruction generated for "Stable Diffusion - Image to Prompts" competition by fine-tuned Llama 2.}
\begin{tabular}{p{0.95\linewidth}}
\hline
Data Preprocessing:\\
• Load the training and test data, ensuring that both image and text prompt data are correctly loaded and aligned.\\
• Apply any necessary data preprocessing, including handling missing values, cleaning the data, and addressing class imbalances if present.\\
• Perform image preprocessing, such as resizing and normalizing pixel values.\\
• Tokenize the text prompts using a suitable tokenizer, considering any specific requirements for this task.\\
• Convert tokenized text prompts into numerical representations (embeddings) using a method that captures the semantic meaning of the text effectively.\\
\hline
Model Architecture:\\
• Design a model architecture that combines image and text information effectively. Consider using a combination of convolutional neural networks (CNNs) for image data and recurrent neural networks (RNNs) for text data.\\
• Implement a mechanism for merging or concatenating the image and text features within the model architecture.\\
• Utilize pre-trained embeddings or models if available and relevant to improve model performance.\\
\hline
Model Training and Evaluation:\\
• Split the training data into training and validation sets for model training. Implement a suitable validation strategy, such as k-fold cross-validation, to assess model performance robustly.\\
• Train the model with appropriate hyperparameters, including learning rate, batch size, and optimizer choice. Monitor training with metrics relevant to the task, such as mean cosine similarity.\\
• Implement early stopping to prevent overfitting and save the best-performing model during training.\\
• Evaluate the model using metrics like mean cosine similarity on the validation set to assess its performance.\\
• Consider strategies like learning rate schedules or fine-tuning pre-trained models to improve convergence and model generalization.\\
\hline
\end{tabular}
\label{tab:kaggle1_smart}
\end{table}

\begin{table}[ht]
\centering
\caption{A sample instruction generated for "Regression with a Tabular Paris Housing Price Dataset" competition by Code Llama - Instruct.}
\begin{tabular}{p{0.95\linewidth}}
\hline
As a data analyst, I would follow these steps to solve the regression problem with the tabular Paris housing price dataset:\\ \\
Data Preprocessing:\\
Check for missing values and handle them appropriately.Scale the data using a standard scaler or normalizer to ensure that all features are on the same scale.Split the data into training and testing sets\\
\hline
Model Architecture:\\
Use a linear regression model with a single hidden layer to predict the housing prices.The model should have an input layer with the number of features in the dataset, a hidden layer with a suitable number of neurons, and an output layer with a single neuron.The model should also have an activation function for the hidden layer, such as ReLU or tanh\\
\hline
Model Training:\\
Train the model using the training set and evaluate its performance on the testing set using the RMSE metric.Adjust the hyperparameters of the model, such as the learning rate, number of hidden neurons, and regularization strength, to improve the model's performance.Repeat the training and evaluation process until the model's performance on the testing set is satisfactory.\\
\hline
By following these steps, I can help solve the regression problem with the tabular Paris housing price dataset using a linear regression model with a single hidden layer.\\
\hline
\end{tabular}
\label{tab:kaggle3}
\end{table}

\begin{table}[ht]
\centering
\caption{The best out of top-3 sample instruction generated for "Regression with a Tabular Paris Housing Price Dataset" competition by fine-tuned Llama 2.}
\begin{tabular}{p{0.95\linewidth}}
\hline
Data Preprocessing:\\
-Read the training and testing data from CSV files.\\
-Check for missing values in the dataset and fill them with appropriate values. Perform exploratory data analysis to understand the distribution of variables and any correlations.\\
-Transform categorical variables into numerical codes.\\
-Scale numerical variables using standard scaling techniques.\\
\hline
Model Architecture:\\
-Define the model architecture using a suitable regression algorithm (e.g., Linear Regression, Ridge Regression, Lasso Regression, Elastic Net Regression).\\
-Define the hyperparameters for the model (e.g., regularization strength, learning rate).\\
\hline
Model Training:\\
-Split the data into training and testing sets.\\
-Train the model using the training data.\\
-Evaluate the model using the testing data.\\
-Fine-tune the model by adjusting the hyperparameters.\\
\hline
Model Prediction:\\
-Use the trained model to predict the target variable for the testing data.\\
-Calculate the root mean squared error (RMSE) between the predicted values and the actual values.\\
-Save the predicted values in a CSV file for submission.\\ \\

Note: The code provided is just an example and may need to be modified based on the specific requirements of the task\\
\hline
\end{tabular}
\label{tab:kaggle3_our}
\end{table}

\begin{table}[ht]
\centering
\caption{An automatically chosen and improved with multi-agent LLM best sample instruction generated for "Regression with a Tabular Paris Housing Price Dataset" competition by fine-tuned Llama 2.}
\begin{tabular}{p{0.95\linewidth}}
\hline
Data Preprocessing:\\ 
• Read the training and testing data from CSV files.\\ 
• Check for missing values in the dataset and fill them with appropriate values.\\ 
• Perform exploratory data analysis to understand the distribution of variables and any correlations.\\ 
• Handle categorical variables by transforming them into numerical codes or using one-hot encoding.\\ 
• Scale numerical variables using standard scaling techniques.\\ 
\hline
Model Selection and Training:\\ 
• Choose an appropriate regression algorithm considering the dataset's characteristics. You may consider Linear Regression, Ridge Regression, Lasso Regression, or Elastic Net Regression. The choice should be based on the nature of the problem, linearity assumptions, and potential multicollinearity.\\ 
• Define the hyperparameters for the selected regression model. Pay attention to regularization strength, learning rate, and any other relevant hyperparameters.\\ 
• Split the data into training and testing sets for model evaluation. Additionally, consider using cross-validation for a more robust assessment.\\ 
• Train the selected regression model using the training data. 
• Validate the model assumptions such as linearity, homoscedasticity, and independence of errors.\\ 
• Evaluate the model using the testing data. Calculate the root mean squared error (RMSE) to assess the model's predictive performance.\\ 
• Fine-tune the model by adjusting hyperparameters as needed to improve its performance.\\ 
\hline
\end{tabular}
\label{tab:kaggle3_smart}
\end{table}

\section{Sample code generated by GPT-3.5 using task descriptions and our refined instructions.}
\label{App C}

Tables~\ref{tab:code_comparison1} and~\ref{tab:code_comparison2} represent the examples of code inferred by GPT-3.5 with two variations of task-describing prompts: one with and one without the automatically chosen best instruction. Using a pure task description prompt may result in incomplete ML code generation, lacking an adequately defined model, for example. Conversely, when the task description prompt is enriched with the instruction, GPT-3.5 is driven to produce compilable results~\footnote{Throughout this paper, '***' indicates code segments extracted by the authors.}.

\begin{table}[htbp]
\caption{Code parts generated for Kaggle competition "Stable Diffusion - Image to Prompts" with and without an enriched instruction prompt.}
\hspace{-2.5cm}
\begin{tabular}{|p{0.55\textwidth}|p{0.65\textwidth}|}
\hline
\text{Without instruction} & \text{With instruction} \\
\hline

\begin{lstlisting}[language=Python]

import torch
import torchvision.transforms as transforms
from some_image_to_text_model import ImageToTextModel
from some_text_embedding_model import TextEmbeddingModel
import pandas as pd

# Load your pre-trained models
image_to_text_model = ImageToTextModel.load_pretrained('path_to_image_to_text_model')
text_embedding_model = TextEmbeddingModel.load_pretrained('path_to_text_embedding_model')

# Function to predict prompt from image
def predict_prompt(image_path):
  # Load and preprocess the image
  image = Image.open(image_path)
  transform = transforms.Compose([
    transforms.Resize((224, 224)),
    transforms.ToTensor(),
  ])
  image = transform(image).unsqueeze(0)

  # Use the model to predict the prompt
  predicted_prompt = image_to_text_model.predict(image)
  return predicted_prompt

# Function to convert prompt to embedding
def prompt_to_embedding(prompt):
  # Convert text prompt to embedding
  embedding = text_embedding_model.encode(prompt)
  return embedding

***
\end{lstlisting}
&
\begin{lstlisting}[language=Python]
# Import necessary libraries
***
import tensorflow as tf
from tensorflow.keras.preprocessing.image import ImageDataGenerator
from tensorflow.keras.preprocessing.text import Tokenizer
from tensorflow.keras.layers import Input, Embedding, LSTM, Conv2D, Flatten, concatenate
from tensorflow.keras.models import Model
***

# Data Preprocessing
# Load training and test data
train_data = pd.read_csv('train_data.csv')
test_data = pd.read_csv('test_data.csv')

# Preprocess text prompts
tokenizer = Tokenizer()
tokenizer.fit_on_texts(train_data['prompt'])
train_sequences = tokenizer.texts_to_sequences(train_data['prompt'])
test_sequences = tokenizer.texts_to_sequences(test_data['prompt'])

# Pad sequences to a fixed length if necessary
max_seq_length = 20
train_sequences = tf.keras.preprocessing.sequence.pad_sequences(train_sequences, maxlen=max_seq_length)
test_sequences = tf.keras.preprocessing.sequence.pad_sequences(test_sequences, maxlen=max_seq_length)

# Image preprocessing (resize, normalize)
image_size = (128, 128)
train_datagen = ImageDataGenerator(rescale=1./255, validation_split=0.2)
train_generator = train_datagen.flow_from_dataframe(***)
valid_generator = train_datagen.flow_from_dataframe(***)

# Model Architecture
# Define text input
text_input = Input(shape=(max_seq_length,))
text_embedding = Embedding(input_dim=len(tokenizer.word_index) + 1, output_dim=128)(text_input)
text_lstm = LSTM(128)(text_embedding)

# Define image input
image_input = Input(shape=(image_size[0], image_size[1], 3))
image_cnn = Conv2D(64, (3, 3), activation='relu')(image_input)
image_flatten = Flatten()(image_cnn)

***
\end{lstlisting} \\
\hline
\end{tabular}

\label{tab:code_comparison1}
\end{table}

\begin{table}[htbp]
\caption{Code parts generated for Kaggle competition "Regression with a Tabular Paris Housing Price Dataset" with and without an enriched instruction prompt.}
\hspace{-2.5cm}
\begin{tabular}{|p{0.55\textwidth}|p{0.65\textwidth}|}
\hline
\text{Without instruction} & \text{With instruction} \\
\hline

\begin{lstlisting}[language=Python]

import pandas as pd
from sklearn.model_selection import train_test_split
from sklearn.ensemble import RandomForestRegressor
from sklearn.metrics import mean_squared_error
import numpy as np

# Load the data
train_data = pd.read_csv('train.csv')
test_data = pd.read_csv('test.csv')

# Separate features and target
X = train_data.drop('price', axis=1)
y = train_data['price']

# Split the data into training and validation sets
X_train, X_val, y_train, y_val = train_test_split(X, y, test_size=0.2, random_state=42)

# Initialize the model
model = RandomForestRegressor(random_state=42)

# Train the model
model.fit(X_train, y_train)

# Predict on validation set
val_predictions = model.predict(X_val)

# Calculate RMSE
rmse = np.sqrt(mean_squared_error(y_val, val_predictions))
print(f'Validation RMSE: {rmse}')

# Predict on test data
test_predictions = model.predict(test_data)
***

\end{lstlisting}
&
\begin{lstlisting}[language=Python]
import pandas as pd
import numpy as np
from sklearn.model_selection import train_test_split
from sklearn.preprocessing import StandardScaler
from sklearn.linear_model import LinearRegression, Ridge, Lasso, ElasticNet
from sklearn.metrics import mean_squared_error

# Data Preprocessing
# 1. Read the training and testing data from CSV files.
train_data = pd.read_csv('train.csv')
test_data = pd.read_csv('test.csv')

# 2. Check for missing values in the dataset and fill them with appropriate values.
train_data.fillna(train_data.mean(), inplace=True)
test_data.fillna(test_data.mean(), inplace=True)

***

# 5. Scale numerical variables using standard scaling techniques.
scaler = StandardScaler()
train_data[['squareMeters', 'numberOfRooms', 'floors', 'numPrevOwners', 'made', 'garage']] = \
  scaler.fit_transform(train_data[['squareMeters', 'numberOfRooms', 'floors', 'numPrevOwners', 'made', 'garage']])
test_data*** = \
  scaler.transform(***)

# Model Selection and Training

# 1. Choose an appropriate regression algorithm considering the dataset's characteristics.
# Let's use Ridge Regression as an example.
regression_model = Ridge(alpha=1.0) # You can adjust alpha as needed.

# 3. Split the data into training and testing sets for model evaluation.
X = train_data.drop(columns=['id', 'price'])
y = train_data['price']
X_train, X_test, y_train, y_test = train_test_split(X, y, test_size=0.2, random_state=42)

# 4. Train the selected regression model using the training data.
regression_model.fit(X_train, y_train)

# 5. Evaluate the model using the testing data.
y_pred = regression_model.predict(X_test)
rmse = np.sqrt(mean_squared_error(y_test, y_pred))
print(f'Root Mean Squared Error (RMSE): {rmse}')

# Model Prediction and Submission

# 1. Use the trained model to predict the target variable for the testing data.
test_features = test_data.drop(columns=['id'])
predicted_prices = regression_model.predict(test_features)
***

\end{lstlisting} \\
\hline
\end{tabular}

\label{tab:code_comparison2}
\end{table}

\section{List of the competitions used for validation.}
\label{App D}

This section provides a list of the competitions used to validate our approach. Table \ref{tab:id} presents a mapping between competition IDs and their corresponding names. These competitions, sourced from the Kaggle platform, represent a range of machine learning tasks including regression, binary classification, and specialized problems like image-to-prompt generation.

\begin{table}[t]
\centering
\caption{Competition ID to competition name mapping.}
\begin{tabular}{|c|c|}
\hline
\makecell{Competition \\Id}&\makecell{Competition \\name}\\ \hline
C1&\makecell{"Feature Imputation\\ with a Heat Flux Dataset"\\~\citep{playground-series-s3e15}}\\
C2&\makecell{"Binary Classification of \\Machine Failures"\\~\citep{playground-series-s3e17}}\\
C3&\makecell{Predict CO2 Emissions\ in Rwanda\\~\citep{playground-series-s3e20}}\\
C4&\makecell{CommonLit - Evaluate \\ Student Summaries\\~\citep{commonlit-evaluate-student-summaries}}\\
C5&\makecell{Binary Classification with a \\Tabular Credit Card Fraud \\Dataset~\citep{playground-series-s3e4}}\\
C6&\makecell{Stable Diffusion - Image to \\Prompts~\citep{stable-diffusion-image-to-prompts}}\\
C7&\makecell{Regression with a Tabular\\ California Housing Dataset\\~\citep{playground-series-s3e1}}\\
C8&\makecell{Regression with a\\ Crab Age Dataset\\~\citep{playground-series-s3e16}}\\
C9&\makecell{Regression with a\\ Wild Blueberry Yield Dataset\\~\citep{playground-series-s3e14}}\\
C10&\makecell{Binary Classification with\\ a Bank Churn Dataset\\~\citep{playground-series-s4e1}}\\
\hline
\end{tabular}
\label{tab:id}
\end{table}

\end{document}